\documentclass[10pt,twocolumn,letterpaper]{article}

\usepackage{cvpr}      
\usepackage[ruled]{algorithm2e}

\definecolor{cvprblue}{rgb}{0.21,0.49,0.74}
\usepackage[pagebackref,breaklinks,colorlinks,allcolors=cvprblue]{hyperref}
\usepackage{graphicx} 
\usepackage{float}  
\usepackage{arydshln} 
\usepackage{makecell} 
\usepackage{footmisc}
\usepackage{url}

\usepackage{tikz}
\usepackage{comment}
\usepackage{enumitem}

\usepackage{color}

\usepackage{verbatim}
\usepackage{multirow}
\usepackage{enumitem}
\usepackage{mathrsfs}
\usepackage{pifont}
\usepackage{bm}

\makeatletter
\def\blfootnote{\xdef\@thefnmark{}\@footnotetext}
\makeatother

\def\paperID{4785} 
\def\confName{CVPR}
\def\confYear{2025}

\title{IterIS: Iterative Inference-Solving Alignment for LoRA Merging}

\author{
Hongxu Chen\textsuperscript{1,2} \quad
Zhen Wang\textsuperscript{2} \quad
Runshi Li\textsuperscript{1} \quad
Bowei Zhu\textsuperscript{1} \quad
Long Chen\textsuperscript{2$\dagger$} \\
\textsuperscript{1}University of Science and Technology of China \;
\textsuperscript{2}Hong Kong University of Science and Technology \\
\small\tt \href{https://github.com/HKUST-LongGroup/IterIS-merging}{\textcolor{purple}{https://github.com/HKUST-LongGroup/IterIS-merging}}
}

\begin{document}

\twocolumn[{%
\maketitle
\begin{center}
    \centering
    \captionsetup{type=figure}
     \vspace{-1em}
     \includegraphics[width=0.97\linewidth]{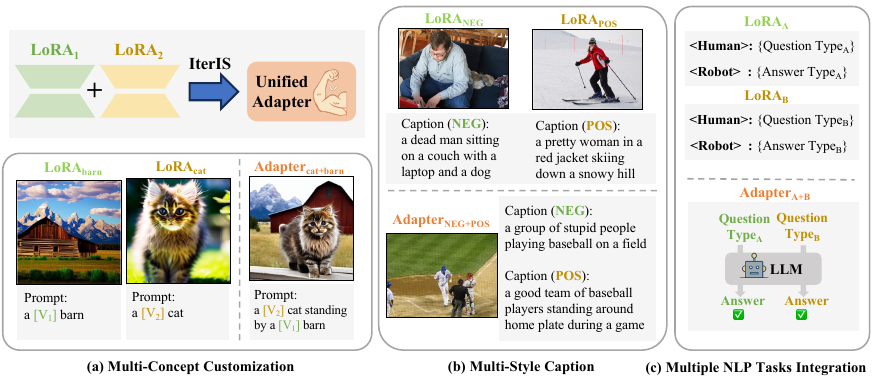}
    \vspace{-1.5em}
    \caption{\textbf{Overview of the application of our method (IterIS) across multiple domains.} Our general method is adaptable for merging LoRAs in various contexts. IterIS can be applied to \textbf{(a)} text-to-image diffusion models for multi-concept customization, \textbf{(b)} vision-language models for multi-style caption generation, and \textbf{(c)} large language models for multiple NLP tasks integration.}
    \label{fig:intro}
\end{center}%
}]

\begin{abstract}
	
Low-rank adaptations (LoRA) are widely used to fine-tune large models across various domains for specific downstream tasks. While task-specific LoRAs are often available, concerns about data privacy and intellectual property can restrict access to training data, limiting the acquisition of a multi-task model through gradient-based training. In response, LoRA merging presents an effective solution by combining multiple LoRAs into a unified adapter while maintaining data privacy. Prior works on LoRA merging primarily frame it as an optimization problem, yet these approaches face several limitations, including the rough assumption about input features utilized in optimization, massive sample requirements, and the unbalanced optimization objective. These limitations can significantly degrade performance. To address these, we propose a novel optimization-based method, named \textbf{IterIS}: 1) We formulate LoRA merging as an advanced optimization problem to mitigate the rough assumption. Additionally, we employ an iterative inference-solving framework in our algorithm. It can progressively refine the optimization objective for improved performance. 2) We introduce an efficient regularization term to reduce the need for massive sample requirements (requiring only 1-5\% of the unlabeled samples compared to prior methods). 3) We utilize adaptive weights in the optimization objective to mitigate potential unbalances in LoRA merging process. 
Our method demonstrates significant improvements over multiple baselines and state-of-the-art methods in composing tasks for text-to-image diffusion, vision-language models, and large language models. Furthermore, our layer-wise algorithm can achieve convergence with minimal steps, ensuring efficiency in both memory and computation.\blfootnote{Work was done when Hongxu Chen and Zhen Wang were visiting students at HKUST. $^\dagger$Long Chen is the corresponding author.}
\end{abstract}

\begin{figure}[t]
    \centering
    \includegraphics[width=0.478\textwidth]{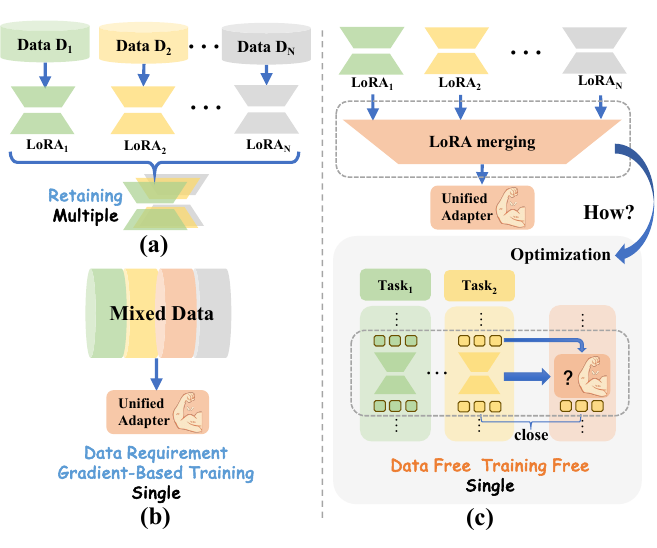} 
    \vspace{-2.2em}
    \caption{\textbf{Overview of methods for multi-task application} \textbf{(a)} Retain all LoRAs fine-tuned on task-specific datasets. \textbf{(b)} Train unified adapters using gradient-based methods on mixed datasets for multi-tasking. \textbf{(c)} Create each unified adapter via LoRA merging without labeled data or gradient-based training. Most methods formulate LoRA merging as an optimization problem to align features and solve for each unified adapter.}
    \label{fig:intro_pic1}
\end{figure}

\section{Introduction}
\label{sec:intro}


Recent advancements have produced increasingly larger pre-trained models with notable performance improvements \cite{rombach2022high, li2022blip, liu2024visual, https://doi.org/10.48550/arxiv.2210.11416, radford2018improving, radford2019language, touvron2023llama}. However, their size makes full fine-tuning resource-intensive. To mitigate this, various parameter-efficient fine-tuning methods~\cite{han2024parameter} have emerged. Especially, low-rank adaptation tuning (LoRA-tuning) \cite{hu2021lora} stands out for its effectiveness. By learning low-rank adaptations within linear layers, LoRA-tuning efficiently captures task-specific information, making it a popular choice for fine-tuning large models. As shown in Figure~\ref{fig:intro}, LoRA-tuning can be effectively applied to various models, including text-to-image diffusion models, vision-language models, and large language models \cite{von-platen-etal-2022-diffusers, zheng2024llamafactory}.




For multi-task applications, a common approach is learning a set of task-specific LoRAs, and each for a specific task. Take the multi-concept customization task as an example (\cf, Figure~\ref{fig:intro}(a)), we learn a single LoRA for each visual concept. Then, these specialized LoRAs are applied separately to manage each task. However, when the specific type of testing task is unknown, the model cannot automatically identify and utilize the corresponding LoRA. Therefore, a straightforward improvement is to develop a single unified adapter, which is capable of managing multiple tasks. As illustrated in Figure~\ref{fig:intro_pic1}(b), we can acquire a unified adapter by learning from the mixed dataset across multiple tasks. However, gradient-based training on large models requires substantial computational resources and time. Moreover, data privacy and intellectual property concerns often limit access to training datasets, further complicating the creation of the unified adapter.

To bypass gradient-based training and preserve data privacy, some \emph{LoRA merging} methods~\cite{gu2024mix, huang2023lorahub, kumari2023multi, jin2022dataless, shah2025ziplora, zhong2024multi, zhang2023composing, prabhakar2024lora} have been proposed. These methods usually formulate LoRA merging as an optimization problem. As shown in Figure~\ref{fig:intro_pic1}(c), they assume all LoRAs share the same pre-trained model, enabling a training-free and layer-wise merging process to acquire each unified adapter.
The most intuitive way is linear merging, \ie, linearly combining the parameters of multiple LoRAs \cite{zhang2023composing, zhong2024multi}. However, the feasibility of it is under the isotropic assumption: \emph{the inputs features to LoRAs follow isotropic distributions}. Since this simplified assumption often does not hold in practice, linear merging generally results in serious performance degradation. 
To move beyond the isotropic assumption, some \emph{real-distribution-based merging} methods are proposed~\cite{jin2022dataless, shah2025ziplora, gu2024mix, kumari2023multi}. These methods formulate the optimization objective utilizing these features acquired from the real distribution. Specifically, they generally involve three steps: \textbf{1)} Extract input features for LoRAs by performing inference on samples. \textbf{2)} Utilize these features to formulate each optimization objective, aiming to align the output features of the unified adapter with those of each LoRA 
(\cf, Figure~\ref{fig:intro_pic1}(c)). \textbf{3)} Acquire each unified adapter using the closed-form solution.

\begin{figure}[t]
    \centering
    \includegraphics[width=0.478\textwidth]{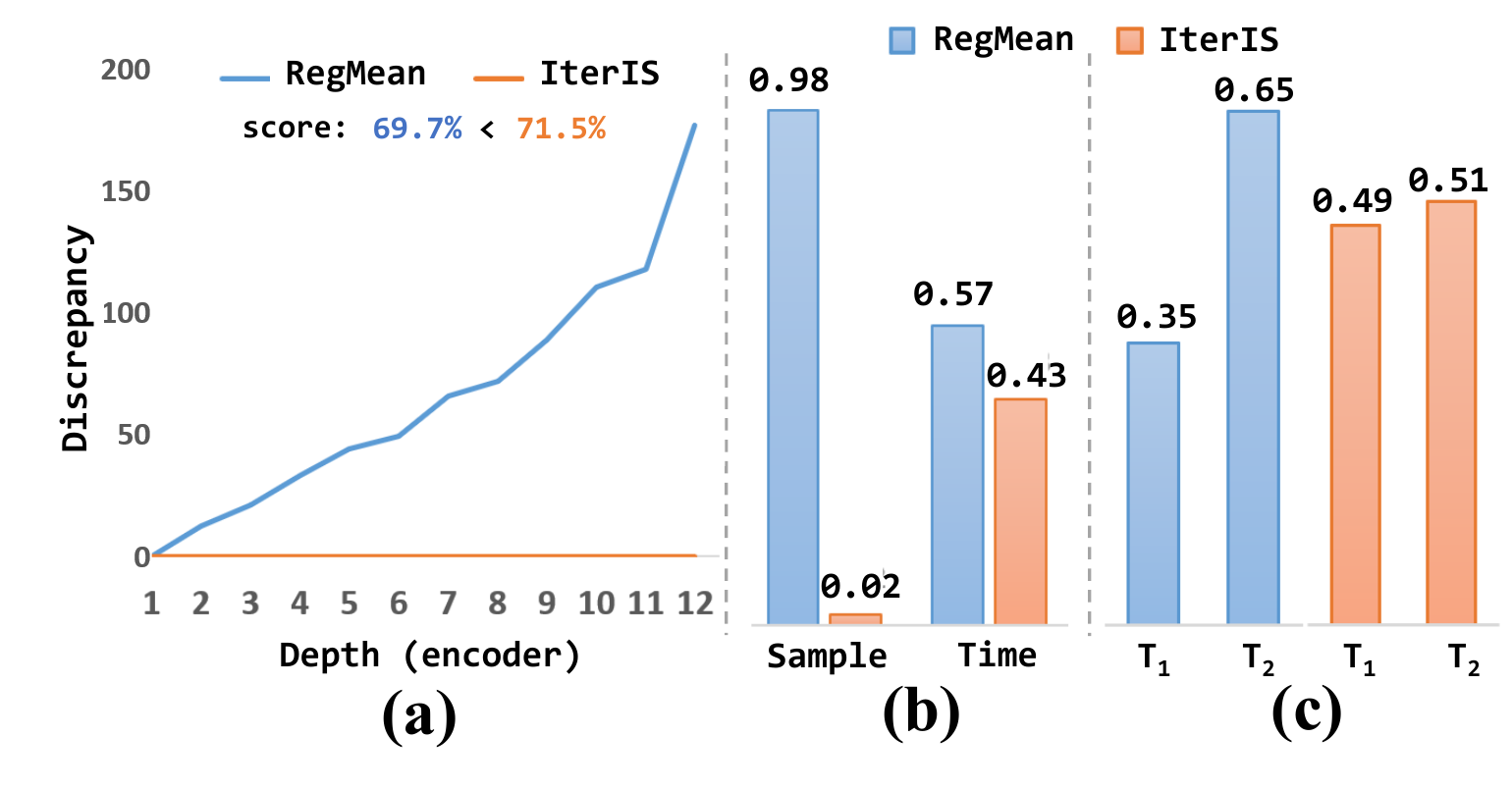}
    \vspace{-2em}
\caption{\textbf{Three key limitations of real-distribution-based merging methods and our improvements.} In the scenario of combining the COLA~\cite{warstadt2019cola} and MNLI~\cite{williams2017broad} tasks in NLP, we compare the representative real-distribution-based merging method, RegMean~\cite{jin2022dataless}, with our proposed method: 
\textbf{(a)} RegMean exhibits increasing discrepancies with deeper encoder layers, while IterIS can fully resolve discrepancies. The value of the ``score'' metric reflects the performance of the method. \textbf{(b)} Comparison of sample requirements and runtime proportions between our method and RegMean. \textbf{(c)} Proportional magnitudes of terms \( T_1 \) and \( T_2 \) in the optimization objective, demonstrating balanced values for our method and imbalance for RegMean.}
\vspace{-1em}
    \label{fig:intro_pic3}
\end{figure}

In this paper, we argue that almost all existing real-distribution-based merging methods still have three overlooked limitations:

\begin{itemize}
    \item \textbf{Rough Assumption\footnote{We use the term ``rough assumption'' to refer to the assumption in this context. \label{rough assumption}}:} They assume the input features for each individual LoRA to be those of the unified adapter. As Figure~\ref{fig:intro_pic3}(a) shows, the discrepancy (measured by Frobenius distance) between approximated and true features increases with encoder layer depth. This rough approximation significantly limits overall performance.
    \item \textbf{Massive Sample Requirement:} They usually require massive samples (\cf, Figure~\ref{fig:intro_pic3}(b)) to enhance algorithm robustness. However, obtaining such a large sample size is often impractical in real-world applications and leads to increased computational time and resource consumption.
    \item \textbf{Unbalanced Optimization:} Significant variations in the magnitudes of output features across task-specific LoRAs can introduce bias into the optimization process as illustrated in Figure~\ref{fig:intro_pic3}(c). It potentially leads to unbalanced performance within tasks.
\end{itemize}

To address these limitations, we propose a novel real-distribution-based merging method, \emph{IterIS}.
To relax the rough assumption\footref{rough assumption}, besides the features for LoRAs, it directly extracts the input features for each \emph{unified adapter}. 
These features are then used to formulate our optimization objective, providing a more accurate representation compared to previous methods.
Additionally, we employ an iterative inference-solving framework to progressively refine the objective. Specifically, in each iteration: (\emph{i}) We extract the input features for each unified adapter through inference. (\emph{ii}) We then update each objective using these features. (\emph{iii}) Finally, we obtain each unified adapter from the closed-form solution for the next iteration. Furthermore, to address the massive sample requirement, we introduce an efficient regularization term in IterIS, which significantly reduces the number of required samples, only needing 1-5\% of the unlabeled samples compared to prior methods. Additionally, to account for potential unbalance in the optimization, adaptive weights are introduced to balance the magnitudes of the terms in the objective.

Benefiting from the directed acyclic structure of deep learning models, IterIS converges with minimal iterations. Besides, its layer-wise updating can ensure both computational and memory efficiency. We evaluate our methods on various domains, demonstrating its capability for LoRA merging. In summary, our \textbf{contributions} are three-fold:
\textbf{1)} We propose IterIS, a novel and versatile LoRA merging algorithm for various domains, supporting applications on text-to-image diffusion, vision-language models, and large language models;
\textbf{2)} We introduce an advanced representation of LoRA merging by employing a progressively refined optimization objective and an iterative inference framework.
\textbf{3)} We empirically demonstrate the effectiveness of IterIS, highlighting its significant improvements in LoRA merging over previous methods.


\section{Related Work}

\begin{figure*}[t]
    \centering
    \includegraphics[width=\textwidth]{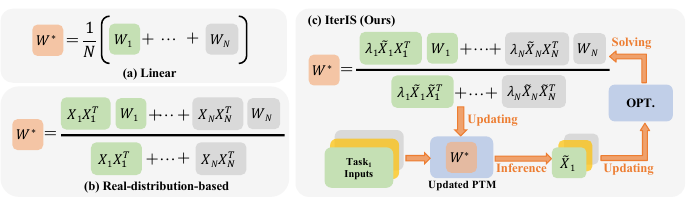} 
    \vspace{-2em}
    \caption{\textbf{Comparison of linear merging, real-distribution-based merging, and IterIS for LoRA Merging.} ``OPT.'' denotes the optimization problem introduced by IterIS. ``PTM'' denotes the pre-trained model. We define $\frac{B}{A}$ to represent $A^{-1}B$. \textbf{(a)} Linear merging combines each individual LoRA linearly. \textbf{(b)} Real-distribution-based merging computes a closed-form solution based on input features for each LoRA. \textbf{(c)} IterIS acquires all the unified adapters through an iterative inference-solving framework.}
    \label{fig:intro_pic2}
\end{figure*}

\noindent \textbf{Parameter-Efficient Fine-Tuning (PEFT)}~\cite{han2024parameter} enables pre-trained models to adapt to downstream tasks with minimal additional parameters, reducing both computational overhead and storage needs. Recent PEFT methods include prompt tuning~\cite{jia2022visual, liu2021p, gal2022image}, adapter tuning~\cite{he2021effectiveness, zhang2023llama}, memory-efficient tuning~\cite{diao2024sherl,diao2024unipt}, and low-rank adaptation tuning (LoRA-tuning)~\cite{hu2021lora}. Prompt tuning introduces a set of task-specific learnable tokens, while adapter tuning adds lightweight modules between layers to capture task-specific knowledge. Among various PEFT methods, LoRA-tuning has become one of the most widely adopted techniques. It has been widely applied for diffusion models and large language models for various downstream tasks~\cite{zheng2024llamafactory, hu2021lora, gu2024mix, zhong2024multi}. By incorporating low-rank decomposition matrices into select layers, LoRA captures task-specific nuances efficiently. Its memory and computational efficiency make it a powerful solution for fine-tuning large models.

\noindent\textbf{LoRA Merging} ~\cite{gu2024mix, huang2023lorahub, kumari2023multi, jin2022dataless, shah2025ziplora, zhong2024multi, zhang2023composing, prabhakar2024lora} is a technique designed to combine multiple LoRAs into a single unified adapter. The simplest approach referred to as linear merging~\cite{zhang2023composing}, involves taking a linear combination of individual LoRAs to
create a unified adapter. This method assumes the inputs to LoRA follow isotropic distributions. However, this assumption is overly simplistic and may lead to performance degradation in real-world applications.

To move beyond the assumption of isotropic distributions, recent studies have proposed LoRA merging as an optimization problem using features from real distribution. The optimization objective aims to align output features between the unified adapter and each individual LoRA~\cite{shah2025ziplora, gu2024mix, kumari2023multi, jin2022dataless}. In the context of text-to-image diffusion, custom diffusion~\cite{liu2019multi} uses constrained optimization to integrate multiple concepts by adjusting key and value projection matrices for text feature alignment. Mix-of-Show~\cite{gu2024mix} adopts a similar optimization strategy during gradient fusion to merge multiple LoRAs, using LBFGS~\cite{liu1989limited} to find optimal solutions.  In integrating multiple NLP tasks, RegMean~\cite{jin2022dataless} pursues the same alignment objective as Mix-of-Show but achieves it through a closed-form solution for greater computational efficiency. These approaches assume that the input features for each individual LoRA to be those for the unified adapter, presenting a rough approximation that can cause serious performance degradation.  In contrast, we directly extract the input features for the unified adapter and refine the optimization objective iteratively, which successfully mitigates the performance degradation.


\section{Method}
In this section, we begin with a brief overview of the LoRA merging's background, followed by an analysis of the limitations of existing methods, which naturally motivates our proposed approach, IterIS.  
\subsection{Preliminaries}

\noindent{\textbf{Definition of LoRA Merging.}}  
Given a pre-trained model and $N$ distinct tasks, we derive $N$ sets of task-specific LoRAs, each fine-tuned on a corresponding dataset and consistently applied at the same positions within the pre-trained model. For each layer, we state that the $N$ task-specific LoRAs operate at the same position, with \( \bm{W}_{i} \) denoting the weights of the \( i \)-th task-specific LoRA. The unified adapter results from merging these $N$ task-specific LoRAs and is denoted by \( \bm{W} \). 
After integrating all unified adapters into the pre-trained model, the resulting model is referred to as the ``unified model''.
It is worth noting that LoRA merging doesn't need any labeled data or gradient-based training.
\noindent{\textbf{Linear Merging.}} 
Linear merging is the most straightforward method. As shown in Figure~\ref{fig:intro_pic2}(a), It merges individual LoRAs by linearly combining their parameters to create a unified adapter. Let \( {\mathcal{\bm{X}}}_i \) denote the random variable representing the input features for the \( i \)-th task, extracted from its corresponding task-specific LoRA. Linear merging can be formulated as the following optimization, aiming to ensure the output of the unified adapter closely matches that of each individual LoRA:
\begin{equation}
\small
    \bm{{W}^{*}} = \mathop{\arg\min}_{\bm{W}} \mathbb{E}_{\mathcal{\bm{X}}} [ \sum_{i=1}^{N} \lambda_i \| \bm{W}_i^{T} {\bm{\mathcal{X}}_i} - \bm{{W}^{T}} {\bm{\mathcal{X}}_i} \|_F^2 ]
    \label{eq:linear}
\end{equation}
where \( \mathbb{E}_\mathcal{\bm{X}}[\cdot] \) denotes the expectation with respect to \( \mathcal{\bm{X}} \) ( \( \mathcal{\bm{X}} \)=\(({\mathcal{\bm{X}}}_1, \dots, {\mathcal{\bm{X}}}_N) \)), and \( \lambda_i \) is a constant. and \( \|\cdot\|_{F} \) represents the Frobenius norm.  It is important to note that Eq.~\eqref{eq:linear} does not have a closed-form solution.

Assuming each \( \bm{\mathcal{X}}_i \) follows an isotropic distribution and \( \bm{\mathcal{X}}_1\)\dots\(\bm{\mathcal{X}}_N \) are mutually independent, Eq.~\eqref{eq:linear} yields a closed-form solution for the unified adapter\footnote{The detailed derivation is left in the appendix. \label{footnote: appendix}}:
\begin{equation}
\small
\bm{W^{*}} = \sum_{i=1}^N \tilde{\lambda}_i \bm{W}_i, \quad \tilde{\lambda}_i = \frac{\lambda_i \mathbb{E}_{\bm{\mathcal{X}}_i}[\|\bm{\mathcal{X}}_i\|_{F}^2]}{\sum_{j=1}^N \lambda_j \mathbb{E}_{\bm{\mathcal{X}}_j}[\|\bm{\mathcal{X}}_j\|_{F}^2]}
\label{eq:linear_solution}
\end{equation}
where $\mathbb{E}_{{\mathcal{X}}_i}[\cdot]$ denotes the expectation with respect to the distribution of $\bm{\mathcal{X}}_i$. This solution represents a linear merging of the parameters of the $N$ task-specific LoRAs. However, assuming an isotropic distribution for each \( \bm{\mathcal{X}}_i \) is \textbf{overly simplistic} and often does not hold in practice, it significantly limits the performance of the unified model. Additionally, treating the weighting coefficients \( \tilde{\lambda_i} \) as hyperparameters, as in PEMs~\cite{zhang2023composing}, greatly expands the hyperparameter space, making manual tuning impractical as the number of tasks ($N$) increases. Therefore, a common and practical approach is to average the $N$ task-specific LoRAs, \ie, $\tilde{\lambda_i}=\frac{1}{N}$.

\noindent{\textbf{Real-distribution-based Merging.}} To relax the assumption of isotropic distribution, some LoRA merging methods utilize features from real distribution~\cite{jin2022dataless, shah2025ziplora, gu2024mix, kumari2023multi}. Specifically, these methods perform inference on unlabeled samples to obtain input features for the LoRAs, and then approximate the mathematical expectation in Eq.~\eqref{eq:linear} using these features. By setting all \(\lambda_i\) to be equal, we can obtain the following optimization\footref{footnote: appendix} from Eq.~\eqref{eq:linear}:
\begin{equation}
\small
	\bm{W}^{*}=\mathop{\arg\min}_{\bm{W}} \textstyle{\sum}_{i=1}^{N}  \| \bm{W}_{i}^{T}\bm{X}_{i} - \bm{W}^{T}\bm{X}_{i}  \|_{F}^{2}
	\label{eq:regmean}
\end{equation}
where \( \bm{X_i} \) represents the input features of the \( i \)-th task samples for the corresponding task-specific LoRA. 
This optimization can be viewed as a linear regression in the matrix space, enabling a closed-form solution\footref{footnote: appendix} for the unified adapter: 
\begin{equation}
\small
	\bm{W}^*=(\sum_{i=1}^N{\bm{X}_i\bm{X}_{i}^{T}})^{-1}(\sum_{i=1}^N{\bm{X}_i\bm{X}_{i}^{T}}\bm{W}_i)
	\label{eq:regmean_solution}
\end{equation}
Building on the derivations above, we highlight three key limitations of exiting real-distribution-based merging methods: \textbf{1)} Eq.~\eqref{eq:regmean} relies on \textbf{rough assumption}\footref{rough assumption}, which assumes the input features for the unified adapter \( \bm{W} \) is identical to \( \bm{X}_i \) (\ie, in Eq.~\eqref{eq:regmean}, \( \bm{X}_i \) in \( \bm{W}^T \bm{X}_i \) represents the approximated input features). Specifically, the input features for the first layer of the unified model are identical to those of the LoRA-tuned models when provided with the same input. However, as model depth increases, the discrepancies grow significantly, making this approximation rough and limiting performance. \textbf{2)} To ensure the invertibility of \(\sum_{i=1}^N \bm{X}_i \bm{X}_i^T\) in Eq.~\eqref{eq:regmean_solution} and enhance robustness, they are usually   accompanied by \textbf{massive sample requirements}. However, this is often impractical in real-world applications. \textbf{3)} Since the weight of \( \bm{W}_i \) in Eq.~\eqref{eq:regmean_solution} is based on the inner product matrix \( \bm{X}_i \bm{X}_i^T \), variations in the magnitudes of \( \bm{X_i} \) can cause the solution to be overly influenced by the most prominent term. It potentially renders Eq.~\eqref{eq:regmean} an \textbf{imbalanced optimization}.

\IncMargin{-0.5em}
\begin{algorithm}[t]
\caption{IterIS (ours)}
\SetKwInOut{Input}{Input}
\SetKwInOut{Output}{Output}
\SetKw{Initial}{Initial:}
\SetKw{keyin}{in}
\Input{
    \hspace*{-0.5em}$\textup{Individual LoRA-tuned models: } f_1, \ldots, f_N$ \\
    \hspace*{-0.3em}$\textup{Number of LoRAs within one model: } J$ \\
    \hspace*{-0.3em}$\textup{Regularization parameter: } \alpha$ \\
    \hspace*{-0.3em}$\textup{Maximum iterations: } \textup{MaxIter}$ \\
    \hspace*{-0.3em}$\textup{Sample inputs across different tasks: } \bm{X}_{1},\ldots,\bm{X}_N$ \\
    \hspace*{-0.3em}$\textup{Input features: } \bm{X}_{nj} \ \forall 1\leq n \leq N, 1 \leq 
 j \leq J$ \\
} 
\Output{$\textup{Merged model: } f_M$ \textup{(\ie, PTM $+$ adapters)}}

\Initial{
    $\bm{W}_{nj} \gets \textup{GetLoraWeights}(f_n, j)$, \\
    \hspace*{3.5em}$\lambda_{nj} \gets \frac{\| \bm{W}_{nj} \|_F^2}{\| \bm{W}_{nj}^T \bm{X}_{nj} \|_F^{2}}$, $\tilde{\bm{X}}_{nj} \gets \bm{X}_{nj}$, \\ 
    \hspace*{3.5em} $\forall 1\leq n \leq  N, 1 \leq j \leq J$\\
}

\For{$\textup{iter} \ \keyin \ 1, 2, \ldots, \textup{MaxIter}$}{
    \For{$j \ \keyin \ 1, 2, \ldots, J$}{
        \For{$n \ \keyin \ 1, 2, \ldots, N$}{
            $\bm{G}_{nj} \gets \tilde{\bm{X}}_{nj} \bm{X}_{nj}^T + \alpha \| \tilde{\bm{X}}_{nj} \bm{X}_{nj}^T \|_F I$\;
            $\tilde{\bm{\bm{G}}}_{nj} \gets \tilde{\bm{X}}_{nj} \tilde{\bm{X}}_{nj}^T + \alpha \| \tilde{\bm{X}}_{nj} \tilde{\bm{X}}_{nj}^T \|_F I$\;
        }
        $\bm{W}^*_j \gets \frac{\sum_{n=1}^N \lambda_{nj} \bm{G}_{nj} \bm{W}_{nj}}{ \sum_{n=1}^N \lambda_{nj} \tilde{\bm{G}}_{nj} }$\;
        \textup{Use} $\bm{W}^*_j$ \textup{to update the $j$-th adapter in} $f_M$\;
    }
    $\{\tilde{\bm{X}}_{nj}\}_{n \in \mathcal{N}, j \in \mathcal{J}} \gets \textup{GetInputFtrs}(f_M, \bm{X}_{1 \ldots N})$\;
}
\Return{$f_M$}
\label{alg:IterIS}

\end{algorithm}
\DecMargin{-0.5em}
\subsection{IterIS Algorithm}
As illustrated in Figure~\ref{fig:intro_pic2}(c), our algorithm employs an iterative {inference-solving framework.} In each iteration, IterIS performs inference using unlabeled samples, extracting input features for all unified adapters. These features are then used to formulate IterIS’s optimization objectives. By solving each optimization, IterIS obtains the unified adapters for the next iteration. This iterative process refines the optimization objective gradually to acquire the unified adapter with increasing performance. Unlike previous methods that usually rely on massive unlabeled samples, IterIS only needs to utilize a small set of unlabeled samples. Additionally, by incorporating an optimization objective with adaptive weights, our method ensures a balanced solution for each unified adapter. We now present a detailed discussion of the three primary improvements introduced in our work.

\noindent\textbf{Iterative Inference-Solving}. We further build upon the core principles of real-distribution-based merging methods. Let \(\tilde{\bm{X}}_{i}\) denote the input features for the unified adapter (\(\bm{W}\)), replacing approximation \(\bm{X}_{i}\). We  derive a more accurate representation for LoRA merging than prior methods: 
\begin{equation}
\small
	\bm{W}^{*} = \mathop{\arg\min}_{\bm{W}} \sum_{i=1}^{N} \lambda_{i} \| \bm{W}_{i}^{T}\bm{X}_{i} - \bm{W}^{T}\tilde{\bm{X}}_{i} \|_{F}^{2}
	\label{eq:IterIS}
\end{equation}
where \(\lambda_i\) is a constant defined later. This optimization objective has the following closed-form solution:
\begin{equation}
\small
	\bm{W}^* = ( \sum_{i=1}^N \lambda_i \tilde{\bm{X}}_i \tilde{\bm{X}}_{i}^{T} )^{-1} ( \sum_{i=1}^N \lambda_i \tilde{\bm{X}}_i \bm{X}_{i}^{T} \bm{W}_i )
	\label{eq:IterIS_solution}
\end{equation}
All \(\tilde{\bm{X}}_{i}\) are updated iteratively, starting with \(\tilde{\bm{X}}_{i} = \bm{X}_{i}\). At each iteration, IterIS has two steps: \textbf{1)} Inference Step: By performing inference on the \(i\)-th task samples within the current unified model, we can extract \(\tilde{\bm{X}}_{i}\). Here, \(\tilde{\bm{X}}_{i}\) represents the input features for the unified adapter,
\textbf{2)} Solving Step: After updating the \(\tilde{\bm{X}}_{i}\) in Eq.~\eqref{eq:IterIS_solution}, we compute the solution \(\bm{W}^{*}\) for each unified adapter. These unified adapters are then incorporated into the pre-trained model, resulting in a new unified model for the next iteration. 
By directly leveraging the input features for the unified adapter, our method mitigates the rough assumption\footref{rough assumption}. Furthermore, through iterative refinement of the optimization objective, IterIS can achieve a more accurate merging representation.
\begin{figure*}[htbp]
    \centering
    \includegraphics[width=\textwidth]{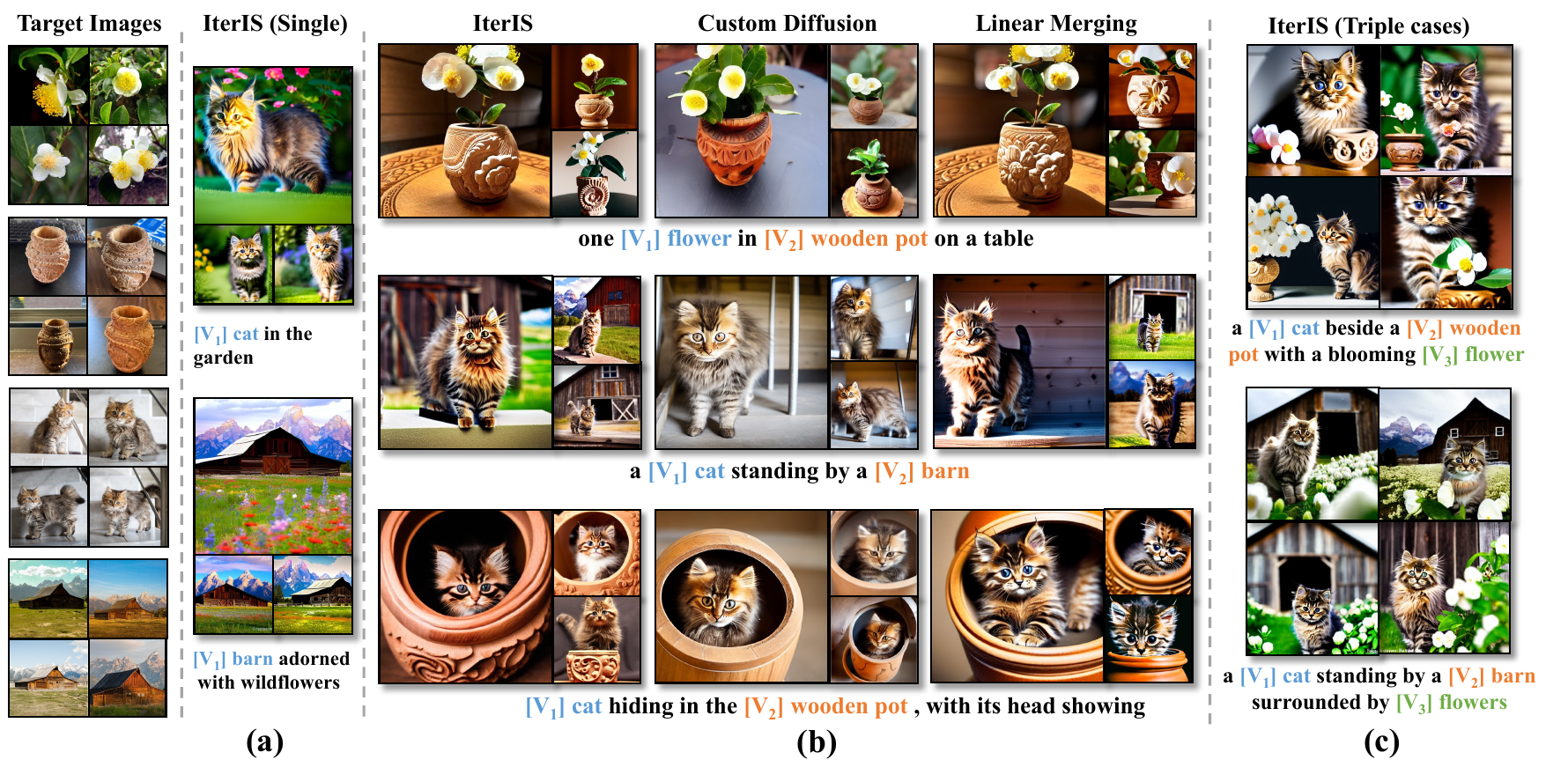}
    \vspace{-2.5em}
    \caption{\textbf{Qualitative results for multi-concept customization.} Target images illustrate single concepts used in composition.
    \textbf{(a)} Single concept generated by IterIS after composing Cat + Barn.
    \textbf{(b)} Comparison of pairwise composition across methods. \textbf{(c)} Triple composition examples generated with IterIS.}
    \label{fig:exp_pic1}
\end{figure*}

\noindent\textbf{Few-Sample Requirement}. Unlike previous methods that require a large number of unlabeled samples, our approach relies on a limited set  
(requiring only 1-5\% of the unlabeled samples compared to prior methods). This is achieved by introducing regularization to the inner product matrices in Eq.~\eqref{eq:IterIS_solution}, replacing them with the following: 
\begin{equation}
\small
\tilde{\bm{X}}_i\tilde{\bm{X}}_i^{T} + \alpha \parallel \tilde{\bm{X}}_i\tilde{\bm{X}}_i^{T} \parallel_F \bm{I}, \tilde{\bm{X}}_i\bm{X}_i^{T} + \alpha \parallel \tilde{\bm{X}}_i\bm{X}_i^{T} \parallel_F \bm{I}
\label{regularization}
\end{equation}
where \(\bm{I}\) denotes the identity matrix. Notably, this regularization alleviates overfitting on these samples, thereby enhancing the algorithm's robustness. In our experiments, we typically set \(\alpha\) to \(1 \times 10^{-4}\) or lower.

\noindent\textbf{Adaptive Weight Balancing}. 
Setting uniform weights for $\lambda_i$ in Eq.~\eqref{eq:IterIS} leads to the same imbalanced issues encountered in previous works. To mitigate this, we introduce adaptive weights for balancing, defined as follows:
\begin{equation}
\small
\lambda _i= \| \bm{W}_i  \| _{F}^{2} \| \bm{W}_{i}^{T}\bm{X}_i  \| _{F}^{-2}
\label{weight_choice}
\end{equation}
This choice of \(\lambda_{i}\) mitigates imbalances in the optimization process, preventing any terms from being overly prioritized.

\noindent\textbf{Analysis of IterIS}. By leveraging the directed acyclic graph structure in deep learning models, we derive an upper bound on the iteration count for our algorithm's convergence\footref{footnote: appendix}. Given an encoder with J-layer self-attention modules, where the Q, K, and V matrices are fine-tuned using LoRA, our algorithm can converge after \(J-1\) iterations.
To prevent overfitting, we cap the iterations at 20 in our experiments. Despite the increased number of iterations, IterIS converges with significantly fewer samples compared to previous methods, maintaining a low overall inference count and ensuring computational efficiency. Additionally, our method bypasses the computational demands of gradient-based training through closed-form solutions and layer-wise updates. Efficiency analysis and full process of IterIS are illustrated in the appendix, with the complete procedure detailed in Algorithm \ref{alg:IterIS}.

\addtolength{\tabcolsep}{-2.5pt}
\begin{table*}[ht]
\fontsize{10}{11}\selectfont
\renewcommand{\arraystretch}{1.1}
\centering
\resizebox{\textwidth}{!}{%
\begin{tabular}{llcccccccc}
\toprule
& \textbf{Method} & \makecell{\textbf{Wooden Pot} \\ \textbf{Teddybear}} & \makecell{\textbf{Flower} \\ \textbf{Dog}} & \makecell{\textbf{Teddybear} \\ \textbf{Tortoise Plushy}} & \makecell{\textbf{Cat} \\ \textbf{Barn}} & \makecell{\textbf{Wooden Pot} \\ \textbf{Cat}} & \makecell{\textbf{Flower} \\ \textbf{Wooden Pot}} & \makecell{\textbf{Dog} \\ \textbf{Sunglasses}} & \textbf{Mean} \\ \midrule
\multirow{4}{*}{\textbf{Image-alignment$_{1}$} }
    & Custom diffusion~\cite{kumari2023multi}  & \textbf{0.570} & 0.603 & \textbf{0.721} & \textbf{0.898} & 0.518 & 0.703 & 0.804 & 0.6881 \\
    & Textual inversion~\cite{gal2022image}      & 0.523 & 0.562 & 0.694 & 0.819 & 0.492 & 0.685 & 0.823 & 0.6569 \\ \cdashline{2-10}[4pt/6pt]
    & Linear merging~\cite{zhang2023composing}   & 0.549 & 0.569 & 0.700 & 0.817 & 0.536 & 0.765 & 0.832 & 0.6811 \\
    & IterIS(Ours) & 0.554 & \textbf{0.609} & 0.700 & 0.820 & \textbf{0.537} & \textbf{0.769} & \textbf{0.833} & \textbf{0.6889} \\ \midrule
\multirow{4}{*}{\textbf{Image-alignment$_{2}$} }
    & Custom diffusion  & 0.772 & 0.768 & 0.683 & 0.587 & \textbf{0.848} & \textbf{0.635} & \textbf{0.671} & 0.7091 \\
    & textual inversion      & 0.793 & 0.778 & 0.677 & 0.588 & 0.751 & 0.610 & 0.624 & 0.6887 \\ \cdashline{2-10}[4pt/6pt]
    & Linear merging   & 0.802 & 0.753 & 0.698 & \textbf{0.617} & 0.755 & 0.606 & 0.643 & 0.6963 \\
    & IterIS(Ours) & \textbf{0.807} & \textbf{0.853} & \textbf{0.702} & 0.615 & 0.759 & 0.608 & 0.643 & \textbf{0.7124} \\ \midrule
\multirow{4}{*}{\textbf{Text-alignment}} 
    & Custom diffusion  & \textbf{0.706} & \textbf{0.710} & 0.510 & 0.542 & 0.661 & 0.643 & \textbf{0.784} & 0.6509 \\
    & Textual inversion       & 0.691 & 0.544 & \textbf{0.614} & 0.579 & 0.641 & 0.631 & 0.754 & 0.6363 \\ \cdashline{2-10}[4pt/6pt]
    & Linear merging   & 0.685 & 0.546 & 0.600 & 0.637 & 0.668 & 0.689 & 0.758 & 0.6547 \\
    & IterIS(Ours) & 0.691 & 0.699 & 0.601 & \textbf{0.639} & \textbf{0.670} & \textbf{0.691} & 0.769 & \textbf{0.6800} \\ 
\bottomrule
\end{tabular}%
}
\vspace{-0.5em}
\caption{\textbf{Quantitative results for multi-concept customization.} ``Image-alignment$_{i}$'' refers to the alignment quality of the \(i\)-th generated object. \textbf{Bold} values indicate the best performance across all methods.}
\vspace{-1em}
\label{table:performance_comparison}
\end{table*}
\addtolength{\tabcolsep}{2.5pt}

\section{Experiments}

In this section, we evaluate the effectiveness of IterIS across various tasks with different pre-trained generative models, including text-to-image diffusion, vision-language models, and large language models. Due to the limited space, more details of the experimental settings and the ablation studies are left in the appendix.
\subsection{IterIS for Text-to-Image Diffusion Model}

\textbf{Experimental Setup.} We applied IterIS to text-to-tmage diffusion model for multi-concept customization, utilizing Stable Diffusion v1.5~\cite{rombach2022high}. Multiple LoRAs were obtained through the widely used Dreambooth-LoRA~\cite{ruiz2023dreambooth} tuning method, with each concept fine-tuned using only 5-10 images without any regularized data. The target images are sourced from the CustomConcept101~\cite{kumari2023multi} and DreamBooth~\cite{ruiz2023dreambooth} datasets. In all experiments, images were processed at a resolution of \(512 \times 512\), employing a DDPM sampler that runs for 100 steps per composition. Each composition features 2-3 concepts, and we set the guidance scale to 12. We utilized 50 input samples for inference to derive input features for our algorithm. Adhering to similar settings as custom diffusion~\cite{kumari2023multi}, we present the results of generating two new concepts within the same scene for the following seven pairs: 1) Wooden Pot+ Teddybear; 2) Flower + Dog; 3) Teddybear + Tortoise Plushy; 4) Cat + Barn; 5) Wooden Pot + Cat; 6) Flower + Wooden Pot; 7) Dog + Sunglasses.

\noindent\textbf{Metrics and Baselines.} Following standard evaluation practices, we assess our method using two key metrics, measured by CLIP-large-patch14~\cite{radford2021learning}: (1) image alignment, which evaluates the coherence of two new concepts within the same scene, and (2) text alignment, which measures the consistency between textual descriptions and their corresponding visual representations. For each composition pair, we generate 400 images across 8 challenging prompts. Each metric is calculated as the average across 400 generated images per composition. We benchmark our approach against linear merging, textual inversion~\cite{gal2022image}, and the state-of-the-art method for multi-concept customization, custom diffusion~\cite{kumari2023multi}, which employs data regularization.

\noindent\textbf{Main Results.} \textbf{1)} Qualitative Evaluation: We present our results on single-concept generation in Figure~\ref{fig:exp_pic1}(a). After composing ``Cat'' and ``Barn'' using our method, the generated concept effectively preserves the majority of the features without any mutual interference. Figure~\ref{fig:exp_pic1}(b) illustrates the generations produced by our method, custom diffusion, and linear merging for pairwise composition. Our approach demonstrates a more effective balance in preserving both concepts. In the case of the Cat + Barn combination, our method successfully retains the features of both the Cat and the Barn, whereas custom diffusion fails to effectively generate the Barn. Figure~\ref{fig:exp_pic1}(c) presents examples of our method applied to triple composition, highlighting its effectiveness in handling combinations involving more concepts.
\textbf{2)} Quantitative Evaluation: Table~\ref{table:performance_comparison} summarizes the results for seven pairwise compositions. Our method consistently outperforms both textual inversion and linear merging almost across all cases. On average, IterIS outperforms linear merging by 0.78\%, 1.61\%, and 2.53\% in image alignment$_1$, image alignment$_2$, and text alignment, respectively. Additionally, IterIS demonstrates competitive performance relative to custom diffusion, slightly surpassing it across all metrics on average. Notably, our method ensures enhanced stability and more balanced metrics, highlighting its robustness. It is worth noting that \emph{IterIS exhibits remarkable adaptability and effectiveness in multi-concept customization, even without specific optimization.} 


\begin{table}[t]
\centering
\fontsize{10}{11}\selectfont
\renewcommand{\arraystretch}{1.3}
\setlength{\tabcolsep}{3pt} 
\resizebox{\columnwidth}{!}{ 
\begin{tabular}{l|cccccc}
\toprule
\textbf{Method} & \textbf{Acc}$_{(pos, neg)}$ & CIDEr & B-1 & B-2 & B-3 & B-4 \\ \midrule
\textbf{POS LoRA} & (0.848, 0.018) & 0.654 & 0.504 & 0.309 & 0.197 & 0.126 \\
\textbf{NEG LoRA} & (0.138, 0.867) & 0.667 & 0.520 & 0.322 & 0.206 & 0.133 \\ 
\textbf{Linear~\cite{zhang2023composing}} & (0.522, 0.557) & 0.752 & 0.535 & 0.338 & 0.218 & 0.142 \\
\textbf{Task arithmetic~\cite{ilharco2022editing}} & (0.522, 0.557) & 0.667 & 0.535 & 0.338 & 0.218 & 0.142 \\ 
\textbf{Ties merging~\cite{yadav2023ties}} & (0.468, 0.543) & 0.739 & 0.519 & 0.330 & 0.214 & 0.141 \\
\textbf{Hyper-linear~\cite{zhang2023composing}} & (0.565, 0.539) & 0.769 & 0.535 & 0.342 & 0.223 & 0.147 \\
\textbf{RegMean~\cite{jin2022dataless}} & (0.624, 0.692) & 0.790 & \textbf{0.558} & 0.355 & 0.231 & 0.150 \\
\textbf{IterIS (Ours)} & \textbf{(0.831, 0.781)} & \textbf{0.794} & 0.551 & \textbf{0.355} & \textbf{0.231} & \textbf{0.152} \\ 
\bottomrule
\end{tabular}
}
\vspace{-0.5em}
\caption{\textbf{Performance comparison for multi-style caption generation.} \textbf{Bold} numbers highlight the best performance.}
\vspace{-2.5em}
\label{tab:metrics_comparison}.
\end{table}

\subsection{IterIS for Vision-Language Model}
\noindent\textbf{Experimental Setup.} We employed IterIS to vision-language models for multi-style caption generation, utilizing BLIP-image-captioning-base~\cite{li2022blip}.  Style-specific LoRAs were fine-tuned on BLIP. We used 50 prompt-image pairs per style-specific captioning task to derive input features. We set \(\alpha = 8 \times 10^{-7}\) and 6 iterations. Our method was evaluated on the combination of two caption text styles. Specifically, we utilized the SentiCap dataset~\cite{mathews2016senticap} to train style-specific LoRAs (``positive" and ``negative").

\begin{table*}[t]
\centering
\fontsize{7}{8.5}\selectfont
\setlength{\tabcolsep}{2.2pt} 
\resizebox{\textwidth}{!}{
\begin{tabular}{l|ccccc|cccccccc}
\toprule
& \multicolumn{5}{c|}{\textbf{In-Domain (Emotion)}} & \multicolumn{8}{c}{\textbf{Multi-Task (GLUE)}} \\ \midrule
\textbf{Method} & Emoint~\cite{mohammad2017wassa} & EC~\cite{ghazi2015detecting} & TEC~\cite{mohammad2012emotional} & ISEAR~\cite{scherer1994evidence} & \textbf{SUM} & MNLI~\cite{williams2017broad} & COLA~\cite{warstadt2019cola} & QQP & QNLI~\cite{rajpurkar2016squad} & MRPC~\cite{dolan2005automatically} & SST2~\cite{socher2013recursive} & RTE~\cite{giampiccolo2007third} & \textbf{SUM} \\ \midrule
\textbf{Linear~\cite{zhang2023composing}} & 0.724 & 0.884 & 0.491 & 0.594 & 2.693 & 0.731 & \textbf{0.678} & 0.827 & 0.890 & 0.781 & 0.932 & 0.786 & 5.625 \\
\textbf{RegMean~\cite{jin2022dataless}} & 0.759 & 0.921 & 0.544 & 0.630 & 2.854 & 0.790 & 0.652 & 0.851 & 0.916 & 0.809 & 0.945 & 0.801 & 5.764\\
\textbf{IterIS (Ours)} & \textbf{0.795} & \textbf{0.942} & \textbf{0.547} & \textbf{0.656} &  \textbf{2.940} & \textbf{0.807} & 0.666 & \textbf{0.853} & \textbf{0.917} & \textbf{0.818} & \textbf{0.946} & \textbf{0.804} & \textbf{5.811} \\\bottomrule
\end{tabular}
}
\vspace{-0.5em}
\caption{\textbf{Performance comparison for in-domain and multi-task integration in NLP.} \textbf{Bold} numbers indicate the best performance.}
\vspace{-1em}
\label{tab:task_integration_comparison}
\end{table*}

\noindent\textbf{Metrics and Baselines.} Following common practices~\cite{klein2021diverse}, we evaluate the performance of the style caption generation using style accuracy\cite{sebastiani2006sentiwordnet}, CIDEr~\cite{vedantam2015cider}, and BLEU 1-4~\cite{papineni2002bleu}. Apart from style accuracy, we report the average of all other metrics across test sets for both styles. Our method is benchmarked against linear merging, RegMean~\cite{jin2022dataless}, task arithmetic~\cite{ilharco2022editing}, ties merging~\cite{yadav2023ties}, and hyper-linear~\cite{zhang2023composing}.

\begin{figure}[t]
    \centering
    \includegraphics[width=0.475\textwidth]{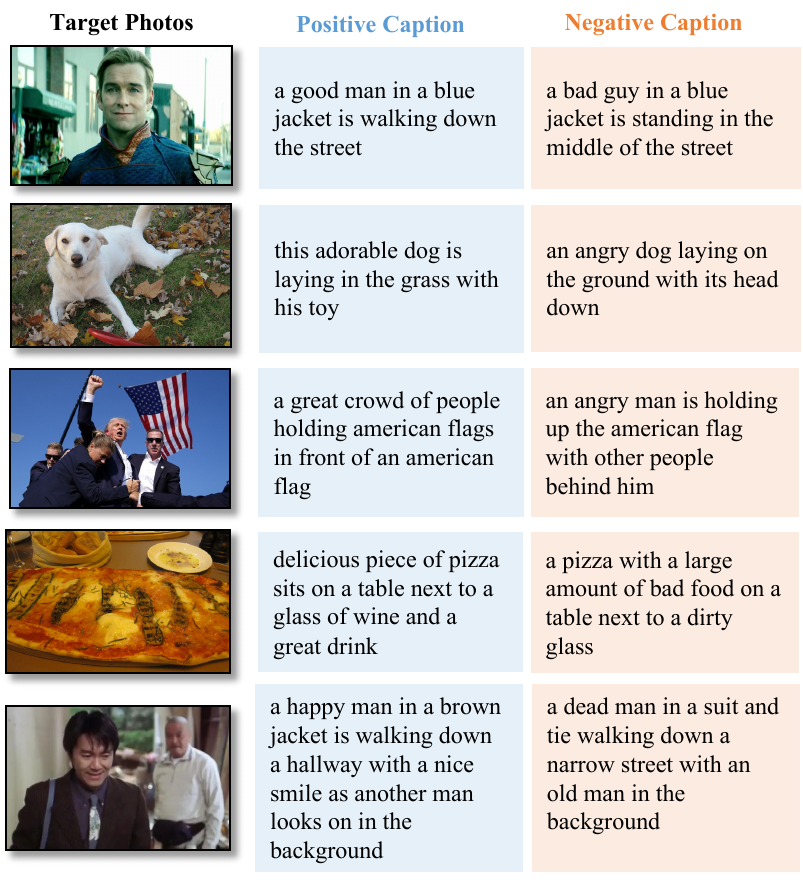}
    \caption{\textbf{Examples of style caption generated by IterIS.} }
    \vspace{-0.5em}
    \label{fig:exp_pic2}
    \vspace{-1em}
\end{figure}

\noindent\textbf{Main Results.} In Table~\ref{tab:metrics_comparison}, we present experimental results for the ``positive'' + ``negative'' style combination. We fine-tuned two specialized LoRA models: POS LoRA, focused on generating positive (POS) captions, and NEG LoRA, specialized in negative (NEG) captions. In single-style generation tests, each LoRA model performs well within its designated style but struggles to produce the other style, with POS LoRA achieving only 1.8\% accuracy in NEG caption generation. Linear merging enables dual-style generation, though it results in limited style controllability, yielding only 52.2\% accuracy for POS and 55.7\% for NEG. RegMean improves style control, achieving 62.4\% accuracy for POS and 69.2\% for NEG captions. Our proposed method achieves the highest caption generation quality, outperforming all other methods and exceeding RegMean by 20.7\% in POS accuracy and 8.9\% in NEG accuracy. Figure~\ref{fig:exp_pic2} shows examples of caption generation in POS and NEG styles utilizing our method. The unified adapters acquired by our method effectively control the generation style while maintaining both caption generation quality and diversity. Additional examples and detailed results are in the appendix.

\subsection{IterIS for Large Language Model}

\noindent\textbf{Experimental Setup.} We applied IterIS to large language models for integrating multiple NLP tasks including in-domain task integration and multi-task integration \footnote{In-domain task: Tasks within the same domain or subject area, often with varied objectives but sharing a common context.
Multi-task: Distinct tasks with different goals, inputs, or outputs, potentially spanning across multiple domains.}. For in-domain task integration, we utilized FLAN-t5-large~\cite{https://doi.org/10.48550/arxiv.2210.11416} on Emotion datasets~\cite{scherer1994evidence, mohammad2012emotional, mohammad2017wassa, mohammad2017wassa} which cover a wide range of sentiments. We applied our method with 50 inputs per task for inference, \(\alpha\) set to \(1 \times 10^{-7}\), and used 10 iterations. We explored all the possible task combinations, specifically two-task, three-task, and four-task integrations for in-domain scenarios. For multi-task integration, we used FLAN-t5-base~\cite{https://doi.org/10.48550/arxiv.2210.11416} as the backbone model across seven datasets in the GLUE benchmark~\cite{wang2018glue}. Input features were derived by performing inference on 50 inputs per task, with \(\alpha\) set to \(1 \times 10^{-4}\) and 10 iterations. We framed classification 
 tasks as generative tasks.

\noindent\textbf{Metrics and Baselines.} For in-domain task integration, we evaluate the performance of the unified model across all task combinations using the F1 score. For multi-task integration, 
Accuracy is used for all the tasks except COLA~\cite{warstadt2019cola}, which is assessed using the Matthews correlation coefficient (MCC)~\cite{matthews1975comparison}. To align the MCC metric range with accuracy, we apply linear normalization to scale MCC values between 0 and 1. Our algorithm is benchmarked against linear merging and the state-of-the-art method, RegMean~\cite{jin2022dataless}. We additionally introduce three extra baselines for the multi-task integration subset, with results provided in the appendix.

\noindent\textbf{Main results.} \textbf{1)} For in-domain task integration, there are a total of 4 in-domain tasks, resulting in $C_{4}^{2} + C_{4}^{3} + C_{4}^{4} = 11$ combinations. We present the average F1 scores across all 11 combinations for each task in the In-Domain section of Table \ref{tab:task_integration_comparison}. It can be observed that our IterIS significantly outperforms both the Linear and RegMean approaches, with improvements over RegMean of 3.6\%, 2.1\%, 0.3\%, and 2.6\% on all the in-domain tasks, respectively. \textbf{2)} For Multi-Task integration, we tested all pairwise combinations, totaling $C_{7}^{2} = 21$ combinations. We calculated the average metric for each GLUE sub-task across these 21 combinations, as shown in the Multi-Task section of Table \ref{tab:task_integration_comparison}. Our method consistently outperforms both Linear and RegMean, with the only exception being a slightly lower average score than linear merging on COLA. Notably, our method shows significant improvement over RegMean on MNLI, COLA, and MRPC, with increases of 1.7\%, 1.4\%, and 0.9\%, respectively. Please refer to the appendix for detailed results.

\section{Conclusion}
In this paper, we propose a novel, versatile algorithm for efficiently merging multiple LoRAs. Building on an iterative inference-solving framework, our method gradually refines the optimization objective, yielding a more effective LoRA merging representation than previous approaches. Additionally, our layer-wise algorithm converges in minimal steps, ensuring efficiency in both memory and computation. Notably, our method outperforms multiple baselines across various domains. Moving forward, we aim to: \textbf{1)} Extend and refine our algorithm to accommodate other PEFT methods, and \textbf{2)} Explore a new perspective on the formulation of the optimization for LoRA merging.

\noindent\textbf{Acknowledgment.}
This work was supported by the Hong Kong SAR RGC Early Career Scheme (26208924), the National Natural Science Foundation of China Young Scholar Fund (62402408), Huawei Gift Fund, the HKUST Sports Science and Technology Research Grant (SSTRG24EG04), the School of the Gifted Young (USTC), and USTC Fellowship for Undergraduate Programs. In particular, Hongxu Chen (First Author) sincerely expresses his gratitude to his parents and his friends Yanghao Wang, Jiazhen Liu, and Jinyue Jiang for their unwavering support.


{
    \small
    \bibliographystyle{ieeenat_fullname}
    \bibliography{main}

\begin{thebibliography}{55}
\providecommand{\natexlab}[1]{#1}
\providecommand{\url}[1]{\texttt{#1}}
\expandafter\ifx\csname urlstyle\endcsname\relax
  \providecommand{\doi}[1]{doi: #1}\else
  \providecommand{\doi}{doi: \begingroup \urlstyle{rm}\Url}\fi

\bibitem[Bostan and Klinger(2018)]{bostan2018analysis}
Laura Ana~Maria Bostan and Roman Klinger.
\newblock An analysis of annotated corpora for emotion classification in text.
\newblock 2018.

\bibitem[Chung et~al.(2022)Chung, Hou, Longpre, Zoph, Tay, Fedus, Li, Wang, Dehghani, Brahma, Webson, Gu, Dai, Suzgun, Chen, Chowdhery, Narang, Mishra, Yu, Zhao, Huang, Dai, Yu, Petrov, Chi, Dean, Devlin, Roberts, Zhou, Le, and Wei]{https://doi.org/10.48550/arxiv.2210.11416}
Hyung~Won Chung, Le Hou, Shayne Longpre, Barret Zoph, Yi Tay, William Fedus, Eric Li, Xuezhi Wang, Mostafa Dehghani, Siddhartha Brahma, Albert Webson, Shixiang~Shane Gu, Zhuyun Dai, Mirac Suzgun, Xinyun Chen, Aakanksha Chowdhery, Sharan Narang, Gaurav Mishra, Adams Yu, Vincent Zhao, Yanping Huang, Andrew Dai, Hongkun Yu, Slav Petrov, Ed~H. Chi, Jeff Dean, Jacob Devlin, Adam Roberts, Denny Zhou, Quoc~V. Le, and Jason Wei.
\newblock Scaling instruction-finetuned language models.
\newblock In \emph{arXiv}, 2022.

\bibitem[Diao et~al.(2024{\natexlab{a}})Diao, Wan, Jia, Zhuge, Zhang, Lu, and Chen]{diao2024sherl}
Haiwen Diao, Bo Wan, Xu Jia, Yunzhi Zhuge, Ying Zhang, Huchuan Lu, and Long Chen.
\newblock Sherl: Synthesizing high accuracy and efficient memory for resource-limited transfer learning.
\newblock In \emph{ECCV}, pages 75--95, 2024{\natexlab{a}}.

\bibitem[Diao et~al.(2024{\natexlab{b}})Diao, Wan, Zhang, Jia, Lu, and Chen]{diao2024unipt}
Haiwen Diao, Bo Wan, Ying Zhang, Xu Jia, Huchuan Lu, and Long Chen.
\newblock Unipt: Universal parallel tuning for transfer learning with efficient parameter and memory.
\newblock In \emph{CVPR}, pages 28729--28740, 2024{\natexlab{b}}.

\bibitem[Dolan and Brockett(2005)]{dolan2005automatically}
Bill Dolan and Chris Brockett.
\newblock Automatically constructing a corpus of sentential paraphrases.
\newblock In \emph{IWP2005}, 2005.

\bibitem[Gal et~al.(2022)Gal, Alaluf, Atzmon, Patashnik, Bermano, Chechik, and Cohen-Or]{gal2022image}
Rinon Gal, Yuval Alaluf, Yuval Atzmon, Or Patashnik, Amit~H Bermano, Gal Chechik, and Daniel Cohen-Or.
\newblock An image is worth one word: Personalizing text-to-image generation using textual inversion.
\newblock \emph{arXiv preprint arXiv:2208.01618}, 2022.

\bibitem[Gan et~al.(2017)Gan, Gan, He, Gao, and Deng]{gan2017stylenet}
Chuang Gan, Zhe Gan, Xiaodong He, Jianfeng Gao, and Li Deng.
\newblock Stylenet: Generating attractive visual captions with styles.
\newblock In \emph{CVPR}, pages 3137--3146, 2017.

\bibitem[Ghazi et~al.(2015)Ghazi, Inkpen, and Szpakowicz]{ghazi2015detecting}
Diman Ghazi, Diana Inkpen, and Stan Szpakowicz.
\newblock Detecting emotion stimuli in emotion-bearing sentences.
\newblock In \emph{CICLing}, pages 152--165. Springer, 2015.

\bibitem[Giampiccolo et~al.(2007)Giampiccolo, Magnini, Dagan, and Dolan]{giampiccolo2007third}
Danilo Giampiccolo, Bernardo Magnini, Ido Dagan, and William~B Dolan.
\newblock The third pascal recognizing textual entailment challenge.
\newblock In \emph{ACL-PASCAL}, pages 1--9, 2007.

\bibitem[Gu et~al.(2024)Gu, Wang, Wu, Shi, Chen, Fan, Xiao, Zhao, Chang, Wu, et~al.]{gu2024mix}
Yuchao Gu, Xintao Wang, Jay~Zhangjie Wu, Yujun Shi, Yunpeng Chen, Zihan Fan, Wuyou Xiao, Rui Zhao, Shuning Chang, Weijia Wu, et~al.
\newblock Mix-of-show: Decentralized low-rank adaptation for multi-concept customization of diffusion models.
\newblock \emph{NeurIPS}, 36, 2024.

\bibitem[Han et~al.(2024)Han, Gao, Liu, Zhang, and Zhang]{han2024parameter}
Zeyu Han, Chao Gao, Jinyang Liu, Jeff Zhang, and Sai~Qian Zhang.
\newblock Parameter-efficient fine-tuning for large models: A comprehensive survey.
\newblock \emph{arXiv preprint arXiv:2403.14608}, 2024.

\bibitem[He et~al.(2016)He, Zhang, Ren, and Sun]{he2016deep}
Kaiming He, Xiangyu Zhang, Shaoqing Ren, and Jian Sun.
\newblock Deep residual learning for image recognition.
\newblock In \emph{CVPR}, pages 770--778, 2016.

\bibitem[He et~al.(2021)He, Liu, Ye, Tan, Ding, Cheng, Low, Bing, and Si]{he2021effectiveness}
Ruidan He, Linlin Liu, Hai Ye, Qingyu Tan, Bosheng Ding, Liying Cheng, Jia-Wei Low, Lidong Bing, and Luo Si.
\newblock On the effectiveness of adapter-based tuning for pretrained language model adaptation.
\newblock \emph{arXiv preprint arXiv:2106.03164}, 2021.

\bibitem[Hu et~al.(2021)Hu, Shen, Wallis, Allen-Zhu, Li, Wang, Wang, and Chen]{hu2021lora}
Edward~J Hu, Yelong Shen, Phillip Wallis, Zeyuan Allen-Zhu, Yuanzhi Li, Shean Wang, Lu Wang, and Weizhu Chen.
\newblock Lora: Low-rank adaptation of large language models.
\newblock \emph{arXiv preprint arXiv:2106.09685}, 2021.

\bibitem[Huang et~al.(2023)Huang, Liu, Lin, Pang, Du, and Lin]{huang2023lorahub}
Chengsong Huang, Qian Liu, Bill~Yuchen Lin, Tianyu Pang, Chao Du, and Min Lin.
\newblock Lorahub: Efficient cross-task generalization via dynamic lora composition.
\newblock \emph{arXiv preprint arXiv:2307.13269}, 2023.

\bibitem[Ilharco et~al.(2022)Ilharco, Ribeiro, Wortsman, Gururangan, Schmidt, Hajishirzi, and Farhadi]{ilharco2022editing}
Gabriel Ilharco, Marco~Tulio Ribeiro, Mitchell Wortsman, Suchin Gururangan, Ludwig Schmidt, Hannaneh Hajishirzi, and Ali Farhadi.
\newblock Editing models with task arithmetic.
\newblock \emph{arXiv preprint arXiv:2212.04089}, 2022.

\bibitem[Jia et~al.(2022)Jia, Tang, Chen, Cardie, Belongie, Hariharan, and Lim]{jia2022visual}
Menglin Jia, Luming Tang, Bor-Chun Chen, Claire Cardie, Serge Belongie, Bharath Hariharan, and Ser-Nam Lim.
\newblock Visual prompt tuning.
\newblock In \emph{ECCV}, pages 709--727. Springer, 2022.

\bibitem[Jin et~al.(2022)Jin, Ren, Preotiuc-Pietro, and Cheng]{jin2022dataless}
Xisen Jin, Xiang Ren, Daniel Preotiuc-Pietro, and Pengxiang Cheng.
\newblock Dataless knowledge fusion by merging weights of language models.
\newblock \emph{arXiv preprint arXiv:2212.09849}, 2022.

\bibitem[Kingma(2014)]{kingma2014adam}
Diederik~P Kingma.
\newblock Adam: A method for stochastic optimization.
\newblock \emph{arXiv preprint arXiv:1412.6980}, 2014.

\bibitem[Klein et~al.(2021)Klein, Mahajan, and Roth]{klein2021diverse}
Franz Klein, Shweta Mahajan, and Stefan Roth.
\newblock Diverse image captioning with grounded style.
\newblock In \emph{DAGM German Conference on Pattern Recognition}, pages 421--436. Springer, 2021.

\bibitem[Kumari et~al.(2023)Kumari, Zhang, Zhang, Shechtman, and Zhu]{kumari2023multi}
Nupur Kumari, Bingliang Zhang, Richard Zhang, Eli Shechtman, and Jun-Yan Zhu.
\newblock Multi-concept customization of text-to-image diffusion.
\newblock In \emph{CVPR}, pages 1931--1941, 2023.

\bibitem[Li et~al.(2022)Li, Li, Xiong, and Hoi]{li2022blip}
Junnan Li, Dongxu Li, Caiming Xiong, and Steven Hoi.
\newblock Blip: Bootstrapping language-image pre-training for unified vision-language understanding and generation.
\newblock In \emph{ICML}, pages 12888--12900. PMLR, 2022.

\bibitem[Liu and Nocedal(1989)]{liu1989limited}
Dong~C Liu and Jorge Nocedal.
\newblock On the limited memory bfgs method for large scale optimization.
\newblock \emph{Mathematical programming}, 45\penalty0 (1):\penalty0 503--528, 1989.

\bibitem[Liu et~al.(2024)Liu, Li, Wu, and Lee]{liu2024visual}
Haotian Liu, Chunyuan Li, Qingyang Wu, and Yong~Jae Lee.
\newblock Visual instruction tuning.
\newblock \emph{NeurIPS}, 36, 2024.

\bibitem[Liu et~al.(2019)Liu, He, Chen, and Gao]{liu2019multi}
Xiaodong Liu, Pengcheng He, Weizhu Chen, and Jianfeng Gao.
\newblock Multi-task deep neural networks for natural language understanding.
\newblock \emph{arXiv preprint arXiv:1901.11504}, 2019.

\bibitem[Liu et~al.(2021)Liu, Ji, Fu, Tam, Du, Yang, and Tang]{liu2021p}
Xiao Liu, Kaixuan Ji, Yicheng Fu, Weng~Lam Tam, Zhengxiao Du, Zhilin Yang, and Jie Tang.
\newblock P-tuning v2: Prompt tuning can be comparable to fine-tuning universally across scales and tasks.
\newblock \emph{arXiv preprint arXiv:2110.07602}, 2021.

\bibitem[Mathews et~al.(2016)Mathews, Xie, and He]{mathews2016senticap}
Alexander Mathews, Lexing Xie, and Xuming He.
\newblock Senticap: Generating image descriptions with sentiments.
\newblock In \emph{AAAI}, 2016.

\bibitem[Matthews(1975)]{matthews1975comparison}
Brian~W Matthews.
\newblock Comparison of the predicted and observed secondary structure of t4 phage lysozyme.
\newblock \emph{Biochimica et Biophysica Acta (BBA)-Protein Structure}, 405\penalty0 (2):\penalty0 442--451, 1975.

\bibitem[Mohammad(2012)]{mohammad2012emotional}
Saif Mohammad.
\newblock \# emotional tweets.
\newblock In \emph{SEM}, pages 246--255, 2012.

\bibitem[Mohammad and Bravo-Marquez(2017)]{mohammad2017wassa}
Saif~M Mohammad and Felipe Bravo-Marquez.
\newblock Wassa-2017 shared task on emotion intensity.
\newblock \emph{arXiv preprint arXiv:1708.03700}, 2017.

\bibitem[Papineni et~al.(2002)Papineni, Roukos, Ward, and Zhu]{papineni2002bleu}
Kishore Papineni, Salim Roukos, Todd Ward, and Wei-Jing Zhu.
\newblock Bleu: a method for automatic evaluation of machine translation.
\newblock In \emph{ACL}, pages 311--318, 2002.

\bibitem[Plummer et~al.(2015)Plummer, Wang, Cervantes, Caicedo, Hockenmaier, and Lazebnik]{plummer2015flickr30k}
Bryan~A Plummer, Liwei Wang, Chris~M Cervantes, Juan~C Caicedo, Julia Hockenmaier, and Svetlana Lazebnik.
\newblock Flickr30k entities: Collecting region-to-phrase correspondences for richer image-to-sentence models.
\newblock In \emph{ICCV}, pages 2641--2649, 2015.

\bibitem[Prabhakar et~al.(2024)Prabhakar, Li, Narasimhan, Kakade, Malach, and Jelassi]{prabhakar2024lora}
Akshara Prabhakar, Yuanzhi Li, Karthik Narasimhan, Sham Kakade, Eran Malach, and Samy Jelassi.
\newblock Lora soups: Merging loras for practical skill composition tasks.
\newblock \emph{arXiv preprint arXiv:2410.13025}, 2024.

\bibitem[Radford(2018)]{radford2018improving}
Alec Radford.
\newblock Improving language understanding by generative pre-training.
\newblock 2018.

\bibitem[Radford et~al.(2019)Radford, Wu, Child, Luan, Amodei, Sutskever, et~al.]{radford2019language}
Alec Radford, Jeffrey Wu, Rewon Child, David Luan, Dario Amodei, Ilya Sutskever, et~al.
\newblock Language models are unsupervised multitask learners.
\newblock \emph{OpenAI blog}, 1\penalty0 (8):\penalty0 9, 2019.

\bibitem[Radford et~al.(2021)Radford, Kim, Hallacy, Ramesh, Goh, Agarwal, Sastry, Askell, Mishkin, Clark, et~al.]{radford2021learning}
Alec Radford, Jong~Wook Kim, Chris Hallacy, Aditya Ramesh, Gabriel Goh, Sandhini Agarwal, Girish Sastry, Amanda Askell, Pamela Mishkin, Jack Clark, et~al.
\newblock Learning transferable visual models from natural language supervision.
\newblock In \emph{ICML}, pages 8748--8763. PMLR, 2021.

\bibitem[Rajpurkar(2016)]{rajpurkar2016squad}
P Rajpurkar.
\newblock Squad: 100,000+ questions for machine comprehension of text.
\newblock \emph{arXiv preprint arXiv:1606.05250}, 2016.

\bibitem[Rombach et~al.(2022)Rombach, Blattmann, Lorenz, Esser, and Ommer]{rombach2022high}
Robin Rombach, Andreas Blattmann, Dominik Lorenz, Patrick Esser, and Bj{\"o}rn Ommer.
\newblock High-resolution image synthesis with latent diffusion models.
\newblock In \emph{CVPR}, pages 10684--10695, 2022.

\bibitem[Ruiz et~al.(2023)Ruiz, Li, Jampani, Pritch, Rubinstein, and Aberman]{ruiz2023dreambooth}
Nataniel Ruiz, Yuanzhen Li, Varun Jampani, Yael Pritch, Michael Rubinstein, and Kfir Aberman.
\newblock Dreambooth: Fine tuning text-to-image diffusion models for subject-driven generation.
\newblock In \emph{CVPR}, pages 22500--22510, 2023.

\bibitem[Scherer and Wallbott(1994)]{scherer1994evidence}
Klaus~R Scherer and Harald~G Wallbott.
\newblock Evidence for universality and cultural variation of differential emotion response patterning.
\newblock \emph{Journal of personality and social psychology}, 66\penalty0 (2):\penalty0 310, 1994.

\bibitem[Sebastiani and Esuli(2006)]{sebastiani2006sentiwordnet}
Fabrizio Sebastiani and Andrea Esuli.
\newblock Sentiwordnet: A publicly available lexical resource for opinion mining.
\newblock In \emph{ELRA}, pages 417--422, 2006.

\bibitem[Shah et~al.(2025)Shah, Ruiz, Cole, Lu, Lazebnik, Li, and Jampani]{shah2025ziplora}
Viraj Shah, Nataniel Ruiz, Forrester Cole, Erika Lu, Svetlana Lazebnik, Yuanzhen Li, and Varun Jampani.
\newblock Ziplora: Any subject in any style by effectively merging loras.
\newblock In \emph{ECCV}, pages 422--438. Springer, 2025.

\bibitem[Socher et~al.(2013)Socher, Perelygin, Wu, Chuang, Manning, Ng, and Potts]{socher2013recursive}
Richard Socher, Alex Perelygin, Jean Wu, Jason Chuang, Christopher~D Manning, Andrew~Y Ng, and Christopher Potts.
\newblock Recursive deep models for semantic compositionality over a sentiment treebank.
\newblock In \emph{EMNLP}, pages 1631--1642, 2013.

\bibitem[Touvron et~al.(2023)Touvron, Lavril, Izacard, Martinet, Lachaux, Lacroix, Rozi{\`e}re, Goyal, Hambro, Azhar, et~al.]{touvron2023llama}
Hugo Touvron, Thibaut Lavril, Gautier Izacard, Xavier Martinet, Marie-Anne Lachaux, Timoth{\'e}e Lacroix, Baptiste Rozi{\`e}re, Naman Goyal, Eric Hambro, Faisal Azhar, et~al.
\newblock Llama: Open and efficient foundation language models.
\newblock \emph{arXiv preprint arXiv:2302.13971}, 2023.

\bibitem[Vaswani(2017)]{vaswani2017attention}
A Vaswani.
\newblock Attention is all you need.
\newblock \emph{NeurIPS}, 2017.

\bibitem[Vedantam et~al.(2015)Vedantam, Lawrence~Zitnick, and Parikh]{vedantam2015cider}
Ramakrishna Vedantam, C Lawrence~Zitnick, and Devi Parikh.
\newblock Cider: Consensus-based image description evaluation.
\newblock In \emph{CVPR}, pages 4566--4575, 2015.

\bibitem[von Platen et~al.(2022)von Platen, Patil, Lozhkov, Cuenca, Lambert, Rasul, Davaadorj, Nair, Paul, Berman, Xu, Liu, and Wolf]{von-platen-etal-2022-diffusers}
Patrick von Platen, Suraj Patil, Anton Lozhkov, Pedro Cuenca, Nathan Lambert, Kashif Rasul, Mishig Davaadorj, Dhruv Nair, Sayak Paul, William Berman, Yiyi Xu, Steven Liu, and Thomas Wolf.
\newblock Diffusers: State-of-the-art diffusion models.
\newblock \url{https://github.com/huggingface/diffusers}, 2022.

\bibitem[Wang(2018)]{wang2018glue}
Alex Wang.
\newblock Glue: A multi-task benchmark and analysis platform for natural language understanding.
\newblock \emph{arXiv preprint arXiv:1804.07461}, 2018.

\bibitem[Warstadt et~al.(2019)Warstadt, Singh, and Bowman]{warstadt2019cola}
Alex Warstadt, Amanpreet Singh, and Samuel~R Bowman.
\newblock Cola: The corpus of linguistic acceptability (with added annotations).
\newblock 2019.

\bibitem[Williams et~al.(2017)Williams, Nangia, and Bowman]{williams2017broad}
Adina Williams, Nikita Nangia, and Samuel~R Bowman.
\newblock A broad-coverage challenge corpus for sentence understanding through inference.
\newblock \emph{arXiv preprint arXiv:1704.05426}, 2017.

\bibitem[Yadav et~al.(2023)Yadav, Tam, Choshen, Raffel, and Bansal]{yadav2023ties}
Prateek Yadav, Derek Tam, Leshem Choshen, Colin~A Raffel, and Mohit Bansal.
\newblock Ties-merging: Resolving interference when merging models.
\newblock \emph{NeurIPS}, 36:\penalty0 7093--7115, 2023.

\bibitem[Zhang et~al.(2023{\natexlab{a}})Zhang, Liu, He, et~al.]{zhang2023composing}
Jinghan Zhang, Junteng Liu, Junxian He, et~al.
\newblock Composing parameter-efficient modules with arithmetic operation.
\newblock \emph{NeurIPS}, 36:\penalty0 12589--12610, 2023{\natexlab{a}}.

\bibitem[Zhang et~al.(2023{\natexlab{b}})Zhang, Han, Liu, Gao, Zhou, Hu, Yan, Lu, Li, and Qiao]{zhang2023llama}
Renrui Zhang, Jiaming Han, Chris Liu, Peng Gao, Aojun Zhou, Xiangfei Hu, Shilin Yan, Pan Lu, Hongsheng Li, and Yu Qiao.
\newblock Llama-adapter: Efficient fine-tuning of language models with zero-init attention.
\newblock \emph{arXiv preprint arXiv:2303.16199}, 2023{\natexlab{b}}.

\bibitem[Zheng et~al.(2024)Zheng, Zhang, Zhang, Ye, Luo, Feng, and Ma]{zheng2024llamafactory}
Yaowei Zheng, Richong Zhang, Junhao Zhang, Yanhan Ye, Zheyan Luo, Zhangchi Feng, and Yongqiang Ma.
\newblock Llamafactory: Unified efficient fine-tuning of 100+ language models.
\newblock In \emph{ACL}, Bangkok, Thailand, 2024. Association for Computational Linguistics.

\bibitem[Zhong et~al.(2024)Zhong, Shen, Wang, Lu, Jiao, Ouyang, Yu, Han, and Chen]{zhong2024multi}
Ming Zhong, Yelong Shen, Shuohang Wang, Yadong Lu, Yizhu Jiao, Siru Ouyang, Donghan Yu, Jiawei Han, and Weizhu Chen.
\newblock Multi-lora composition for image generation.
\newblock \emph{arXiv preprint arXiv:2402.16843}, 2024.

\end{thebibliography}
}

\clearpage
\appendix 
\renewcommand{\thesection}{\Alph{section}} 
\setcounter{page}{1}
\setcounter{equation}{8}
\setcounter{figure}{6}
\setcounter{table}{3}
\maketitlesupplementary
\noindent{In the supplementary materials, we provide:}
\begin{itemize}
\item  \textbf{Mathematical Proofs} (Section~\ref{sec:A}):  
   We present the mathematical proofs for the equations and properties of our proposed algorithm, as discussed in the main text.
\item \textbf{Datasets and Experimental Setup} (Section~\ref{sec:B}):  
   This section details the datasets employed in our experiments, the LoRA training configuration, and the key settings of the baseline implementations.
\item  \textbf{Ablation Studies} (Section~\ref{sec:C}):  
   We report the results of ablation studies, demonstrating the necessity of the three mechanisms integrated into our algorithm.
\item \textbf{Experimental Results} (Section~\ref{sec:D}):  
   A detailed presentation of the experimental results is provided in this section.
\item \textbf{Workflow and Analysis} (Section~\ref{sec:E}):  
   Finally, we illustrate the complete workflow of IterIS and conduct an analysis of the efficiency and limitations of our approach.
\end{itemize}
\section{Detailed Mathematical Derivation}
\label{sec:A}
\subsection{Linear Merging}
\label{sec:linear_merging_subsec}
Consider the following optimization problem:
\begin{equation}
\small
    \bm{{W}^{*}} = \mathop{\arg\min}_{\bm{W}} \mathbb{E}_{\mathcal{\bm{X}}} [ \sum_{i=1}^{N} \lambda_i \| \bm{W}_i^{T} {\bm{\mathcal{X}}_i} - \bm{{W}^{T}} {\bm{\mathcal{X}}_i} \|_F^2 ]
    \label{eq:linear}
\end{equation}
where \( \mathbb{E}_\mathcal{\bm{X}}[\cdot] \) denotes the expectation with respect to \( \mathcal{\bm{X}} \) ( \( \mathcal{\bm{X}} \)=\(({\mathcal{\bm{X}}}_1, \dots, {\mathcal{\bm{X}}}_N) \)), and \( \lambda_i \) is a constant. and \( \|\cdot\|_{F} \) represents the Frobenius norm.   Assuming each \( \bm{\mathcal{X}}_i \) follows an isotropic distribution and \( \bm{\mathcal{X}}_1\)\dots\(\bm{\mathcal{X}}_N \) are mutually independent, we have:
\begin{equation}
\small
    \mathbb{E}_\mathcal{\bm{X}} [ \| \bm{W}_i^{T} \mathcal{\bm{X}}_i - \bm{W}^{T} \mathcal{\bm{X}}_i \|_{F}^2 ] = \frac{1}{D} \mathbb{E}_{\mathcal{\bm{X}}_i} [\| \mathcal{\bm{X}}_i \|_F^2 ] \| \bm{W}_i - \bm{W} \|_{F}^2
    \label{eq:linear_equ}
\end{equation}
where \( D \) is the dimension of each $\mathcal{\bm{X}}_i$. we have the following based on Eq. \ref{eq:linear_equ}:
\small{
    \begin{align}
    \bm{W}^{*} &= \mathop{\arg\min}_{\bm{W}} \mathbb{E}_{\mathcal{\bm{X}}} [\sum_{i=1}^N \lambda_i \| \bm{W}_i^{T} \mathcal{\bm{X}}_i - \bm{W}^{T} \mathcal{\bm{X}}_i \|_{F}^2 ] \\
    &= \mathop{\arg\min}_{\bm{W}} \sum_{i=1}^N \mathbb{E}_{\mathcal{\bm{X}}_i} [\lambda_i \| \bm{W}_i^{T} \mathcal{\bm{X}}_i - \bm{W}^{T} \mathcal{\bm{X}}_i \|_{F}^2 ] \\
    &= \mathop{\arg\min}_{\bm{W}} \sum_{i=1}^N \frac{\lambda_i}{D} \mathbb{E}_{\mathcal{\bm{X}}_i} \left[\| \mathcal{\bm{X}}_i \|_F^2 \right] \| \bm{W}_i - \bm{W} \|_{F}^2
\label{eq:linear_solution_process}
\end{align}
}
This expression admits a closed-form solution.
\begin{equation}
\bm{W}^{*}=\sum_{i=1}^N{\tilde{\lambda}_i\bm{W}_i},\tilde{\lambda}_i=\frac{\lambda _i \mathbb{E}_{\mathcal{\bm{X}}_i} [  \| \mathcal{\bm{X}}_i  \|_{F} ^2  ]}{\Sigma _{j=1}^{N}\lambda _j \mathbb{E}_{\mathcal{\bm{X}}_j} [  \| \mathcal{\bm{X}}_j  \|_{F} ^2  ]}
\label{eq:linear_solution_app}
\end{equation}

\subsection{Real-distribution-based Merging}
We can derive the following from the linear merging derivation in Subsection~\ref{sec:linear_merging_subsec}, and set each $\lambda_{i}$ to 1:
\begin{equation}
\small
    \bm{W}^{*} = \mathop{\arg\min}_{\bm{W}} \sum_{i=1}^{N} \mathbb{E}_{\bm{\mathcal{X}}_i} [ \| \bm{W}_i^{T} \bm{\mathcal{X}}_i - \bm{W}^{T} \bm{\mathcal{X}}_i \|_F^2 ].
    \label{eq:linear_independent}
\end{equation}
To compute each expectation in Eq.~\ref{eq:linear_independent}, we can sample from the distribution of \( \bm{\mathcal{X}}_i \), and using the law of large numbers, approximate the expectation as:  
\small{
    \begin{align}
    \mathbb{E}_{\bm{\mathcal{X}}_i} [ \| \bm{W}_i^{T} \bm{\mathcal{X}}_i - \bm{W}^{T} \bm{\mathcal{X}}_i \|_F^2 ] 
    &\approx \frac{1}{S} \sum_{s=1}^{S} \| \bm{W}_i^{T} \bm{x}_{is} - \bm{W}^{T} \bm{x}_{is} \|_F^2 \\
    &= \frac{1}{S} \| \bm{W}_i^{T} \bm{X}_i - \bm{W}^{T} \bm{X}_i \|_F^2
    \end{align}
}
where $\bm{x}_{is} \in \text{Sample}(\bm{\mathcal{X}}_i)$ and $\bm{X}_i = (\bm{x}_{i1}, \bm{x}_{i2}, \dots, \bm{x}_{iS})$. Therefore, we can reformulate the optimization objective based on the real distribution:
\begin{equation}
\small
    \bm{W}^{*} = \mathop{\arg\min}_{\bm{W}} \sum_{i=1}^{N} \| \bm{W}_i^{T} \bm{X}_i - \bm{W}^{T} \bm{X}_i \|_F^2.
\end{equation}
In fact, this optimization problem can be viewed as a linear regression problem, where $\bm{X}=[ \bm{X}_1,\bm{X}_2,\cdots ,\bm{X}_N ]^T$ is mapped to $\bm{Y}=[ \bm{W}_{1}^{T}\bm{X}_1,\bm{W}_{2}^{T}\bm{X}_2,\cdots ,\bm{W}_{N}^{T}\bm{X}_N ]^T$. Thus, the problem has a closed-form solution: 
\small{
    \begin{align}
\bm{W}^{*}&=( \bm{X}^T\bm{X} ) ^{-1}\bm{X}^T\bm{Y} \\
&=(\sum_{i=1}^N{\bm{X}_i\bm{X}_{i}^{T}})^{-1}(\sum_{i=1}^N{\bm{X}_i\bm{X}_{i}^{T}}\bm{W}_i)
\end{align}
}
Similarly, for the optimization problem corresponding to our algorithm in Eq. 5 of Section 3, we can also obtain its solution.
\subsection{Maximum Iteration Limit for IterIS}
Starting with the topology of the generative model, we construct the gradient computation graph \( G_{\nabla} \), which is a directed acyclic graph (DAG). To create the new graph \( G_M \), we extract all nodes corresponding to the input features for each LoRA. We follow a specific rule to construct \( G_M \): if there is a directed path from node \( A \) to node \( B \) in \( G_{\nabla} \), we establish a directed edge from \( A \) to \( B \) in \( G_M \). The resulting graph \( G_M \) is also a directed acyclic graph. We will demonstrate that \emph{the length of the longest directed path in \( G_M \), originating from any input node (of which there may be multiple), minus one, provides the maximum number of iterations required for the IterIS’s convergence.}

\noindent\textbf{Proof:} Let \( s \) denote the length of the longest path in \( G_M \) originating from the input node \( O_{\text{in}} \). We define a function \( g : v(G_M) \setminus \{O_{\text{in}}\} \rightarrow \{1, 2, \ldots, s\} \), where \( g(A) \) indicates the length of the longest path from \( O_{\text{in}} \) to node \( A \). We will use mathematical induction to demonstrate that at the \( k \)-th iteration, all nodes in \( g^{-1}(\{1, 2, \ldots, k+1\}) \) and their corresponding updated matrices remain unchanged in subsequent iterations. This observation relies on the fact that if the unified adapter's input \( \tilde{X}_i \) remains constant during iterations, the computed \( W_i \) will also remain constant.

\textbf{1. Base Case ( \( k = 0 \) ):}  
   For the initial iteration, it is clear that \( g^{-1}(1) \) is non-empty and contains only nodes with \( O_{\text{in}} \) as their sole parent. Therefore, \( g^{-1}(1) \) is entirely dependent on the input, indicating that these nodes and their corresponding matrices will remain unchanged in subsequent iterations.

\textbf{2. Inductive Step (Assume true for \( k = k_0 \)):}
   Assume that at iteration \( k = k_0 \), all nodes in \( g^{-1}(\{1, 2, \ldots, k_0 + 1\}) \) and their associated matrices remain constant. Since each node in \( g^{-1}(k_0 + 2) \) has its parent nodes contained within \( g^{-1}(\{1, 2, \ldots, k_0 + 1\}) \), it follows that the nodes in \( g^{-1}(k_0 + 2) \) are reliant on the unchanged nodes in \( g^{-1}(\{1, 2, \ldots, k_0 + 1\}) \). Consequently, these nodes and their corresponding updated matrices will also remain constant in subsequent iterations.

By this reasoning, when \( k = s - 1 \), it holds that \( g^{-1}(\{1, 2, \ldots, s\}) = v(G_M) \setminus \{O_{\text{in}}\} \) will remain unchanged in subsequent iterations. Thus, we prove and compute the upper bound on the maximum number of iterations required for the convergence of the IterIS algorithm.

\begin{figure}[t]
    \centering
    \includegraphics[width=0.478\textwidth]{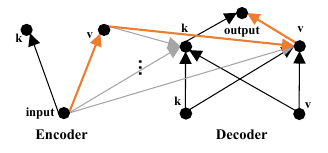} 
    \vspace{-2.2em}
    \caption{\textbf{Illustration of the maximum iteration count in graph \( G_M \).} For a transformer composed of one encoder and one decoder, the abstracted \( G_M \) after LoRA fine-tuning on the \( k \) and \( v \) matrices allows IterIS to converge within two iterations.}
    \label{fig:appendix_1}
\end{figure}

For a transformer model consisting of \( L \) layers of encoders and \( L \) layers of decoders, if we apply LoRA fine-tuning to the \( k \) and \( v \) matrices within the attention modules, our algorithm will converge in no more than \( 3L - 1 \) iterations. As illustrated in Figure~\ref{fig:appendix_1}, the red-marked path has a length of 3, indicating that the algorithm converges after two iterations.

\begin{table}[t]
\centering
\fontsize{12}{13}\selectfont
\renewcommand{\arraystretch}{1.5}
\setlength{\tabcolsep}{5pt}  
\resizebox{\columnwidth}{!}{
\begin{tabular}{l|cccccccc}
\toprule
         & \textbf{B-1} & \textbf{B-2} & \textbf{B-3} & \textbf{B-4} & \textbf{RougeL} & \textbf{CIDEr} & \textbf{ACC$_{POS}$} & \textbf{ACC$_{NEG}$} \\
\midrule
\textbf{LoRA$_{1}$} & 0.570 & 0.369 & 0.241 & 0.159 & 0.426 & 0.808 & 0.848 & 0.018 \\
\textbf{LoRA$_{2}$} & 0.550 & 0.354  & 0.234 & 0.157 & 0.414  & 0.811 & 0.138 & 0.867 \\
\bottomrule
\end{tabular}}
\vspace{-0.7em}
\caption{\textbf{Performance of style-specific LoRAs for multi-style caption on vision-language model.}}
\label{tab:performance_scores_vl}
\end{table}

\begin{table}[t]
\centering
\fontsize{8}{10}\selectfont
\renewcommand{\arraystretch}{1}
\setlength{\tabcolsep}{8pt} 
\resizebox{\columnwidth}{!}{
\begin{tabular}{ccccccc}
\toprule
 \textbf{COLA} & \textbf{RTE} & \textbf{SST2} & \textbf{MNLI} & \textbf{QQP} & \textbf{QNLI} & \textbf{MRPC} \\
\midrule
 {0.532} & {0.812} & {0.946} & {0.855} & {0.858} & {0.920} & {0.831} \\
\bottomrule
\end{tabular}}
\vspace{-1em}
\caption{\textbf{Performance of task-specific LoRAs for multi-task integration on large language model.}}
\label{tab:performance_scores_multi}
\end{table}
\begin{table}[t]
\centering
\fontsize{8}{10}\selectfont  
\renewcommand{\arraystretch}{1}  
\setlength{\tabcolsep}{8pt}  
\begin{tabular}{cccc}
\toprule
 \textbf{TEC} & \textbf{Emoint} & \textbf{ISEAR} & \textbf{EC} \\
\midrule
 0.669 & 0.828 & 0.767 & 0.947 \\
\bottomrule
\end{tabular}
\vspace{-1em}
\caption{\textbf{Performance of task-specific LoRAs for in-domain task integration on large language model.}}
\label{tab:performance_scores_indomain}
\end{table}

\section{Detailed Experimental Setting}
\label{sec:B}
\subsection{Datasets}

\noindent\textbf{DreamBooth~\cite{ruiz2023dreambooth}}. 
The DreamBooth dataset consists of 30 subjects across 15 different classes. Among these, 9 are live subjects (dogs and cats), while the remaining 21 are objects. Each subject is represented by a variable number of images (ranging from 4 to 6), typically captured under diverse conditions, in various environments, and from different angles.

\noindent\textbf{Customconcept101~\cite{kumari2023multi}}. 
The customconcept101 dataset, introduced by custom diffusion~\cite{kumari2023multi}, comprises 101 concepts, each represented by 3 to 15 images, designed to evaluate model customization methods. 

\noindent\textbf{SentiCap}~\cite{mathews2016senticap}. The SentiCap dataset comprises thousands of images with sentiment-labeled captions. The POS subset includes 2,873 positive sentences paired with 998 images for training and 2,019 sentences paired with 673 images for testing. The NEG subset contains 2,468 negative sentences with 997 images for training and 1,509 sentences with 503 images for testing.

\noindent\textbf{GLUE}~\cite{wang2018glue}. The GLUE dataset serves as a benchmark for evaluating natural language understanding models. It encompasses multiple tasks, including sentiment analysis, textual entailment, and sentence similarity. For our evaluation, we utilize the SST-2~\cite{socher2013recursive}, MRPC~\cite{dolan2005automatically}, MNLI~\cite{williams2017broad}, QNLI~\cite{rajpurkar2016squad},  RTE~\cite{giampiccolo2007third}, COLA~\cite{warstadt2019cola}, and QQP  datasets. Evaluations are conducted on the official development sets, as the test labels remain hidden.

\noindent\textbf{Emotion}. For emotion classification, we use the preprocessed datasets provided by Oberlander \& Klinger ~\cite{bostan2018analysis}, specifically TEC~\cite{mohammad2012emotional}, ISEAR~\cite{scherer1994evidence}, Emoint~\cite{mohammad2017wassa}, and Emotion-Cause (EC)~\cite{ghazi2015detecting}, for domain-specific training. All four datasets include the emotion classes of anger, fear, joy, and sadness in their label space, encompassing various scenarios and sources. In our experiments, each dataset is randomly divided into training and testing sets, with five parts for training and one part for testing.

\noindent\textbf{FlickrStyle10K~\cite{gan2017stylenet}}. 
The FlickrStyle10K dataset is derived from the Flickr30K~\cite{plummer2015flickr30k} image caption dataset. It originally contained 10,000 image-caption pairs with stylized captions in humorous and romantic styles. However, only 7,000 pairs from the official training set are currently publicly available.

\begin{table*}[t]
\centering
\fontsize{8.1}{10}\selectfont
\renewcommand{\arraystretch}{1}  
\setlength{\tabcolsep}{5pt}  
\begin{tabular}{c|ccccccccccccccc}
\toprule
\textbf{Iteration} & 0 & 1 & 2 & 3 & 4 & 5 & 6 & 7 & 8 & 9 & 10 & 11 & 12 & 13 & 14 \\
\midrule
\textbf{ACC$_{mean}$} & 0.674 & 0.919 & 0.681 & 0.862 & 0.729 & 0.806 & 0.789 & 0.788 & 0.786 & 0.784 & 0.784 & 0.784 & 0.784 & 0.784 & 0.784 \\
\midrule
\textbf{CIDEr$_{mean}$} & 0.785 & 0.381 & 0.793 & 0.651 & 0.798 & 0.794 & 0.791 & 0.770 & 0.789 & 0.790 & 0.791 & 0.790 & 0.790 & 0.790 & 0.790 \\
\midrule
\textbf{Score$_{aver}$} & 0.730 & 0.650 & 0.737 & 0.756 & 0.764 & 0.800 & 0.790 & 0.779 & 0.788 & 0.787 & 0.788 & 0.787 & 0.787 & 0.787 & 0.787 \\
\bottomrule
\end{tabular}
\vspace{-0.8em}
\caption{\textbf{Performance across different maximum iterations.}}
\vspace{-1.2em}
\label{tab:iteration_scores}
\end{table*}

\subsection{LoRA Training}
\noindent\textbf{LoRAs for Multi-concept Customization}. 
We applied the widely-used DreamBooth-LoRA~\cite{ruiz2023dreambooth} tuning to generate multiple LoRAs for stable diffusion v1.5~\cite{rombach2022high}, with each LoRA dedicated to a single new concept. The LoRA rank was set to 32, and LoRA alpha was set to 1. During the textual inversion phase, we used a learning rate of \( 5 \times 10^{-4} \), with each new concept represented by a single learnable token, all initialized uniformly. In the subsequent fine-tuning phase, the UNet and text encoder were trained with learning rates of \( 10^{-4} \) and \( 10^{-5} \), respectively, while the textual inversion learning rate was maintained at \( 10^{-4} \).  Both phases involved training for 800 steps, with a batch size of 1 and gradient accumulation steps set to 4.

\noindent\textbf{LoRAs for Multi-style Caption}. 
We applied LoRA tuning to fine-tune the BLIP-image-caption-base~\cite{li2022blip}. LoRAs were integrated into all self-attention modules within the text encoder, specifically targeting the \(q\), \(v\), and \(k\) matrices~\cite{vaswani2017attention}. The LoRA rank was set to 32, with \(\alpha\) set to 32, and a dropout rate of 0.05. The training was conducted with a batch size of 8 for a maximum of 15 epochs, using an initial learning rate of 3e-5. The AdamW~\cite{kingma2014adam} optimizer was employed for optimization. Table~\ref{tab:performance_scores_vl} summarizes the key metrics of multiple LoRAs trained in this part.

\begin{figure}[t]
    \centering
    \includegraphics[width=0.478\textwidth]{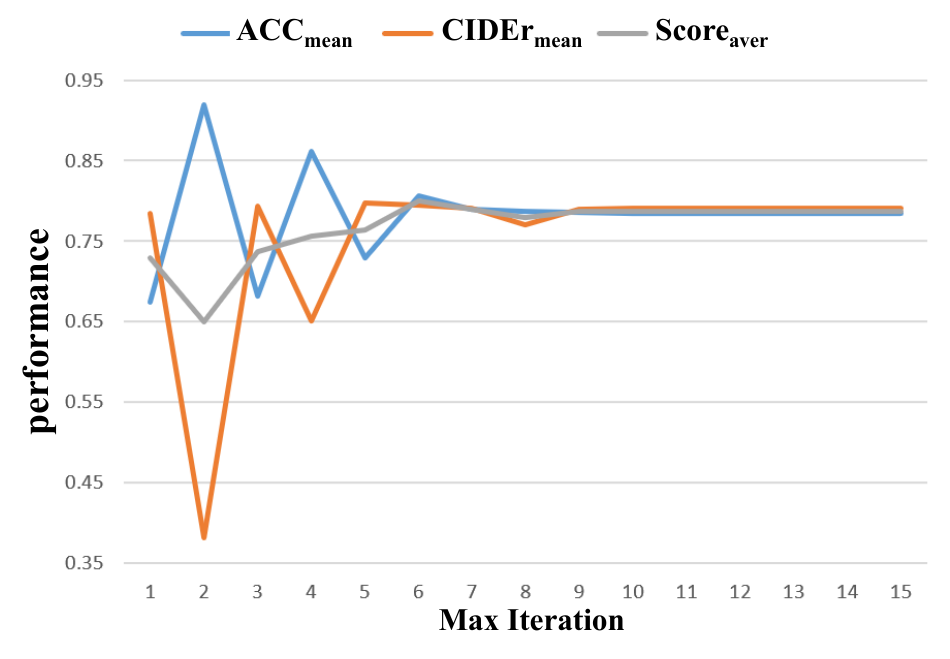} 
    \vspace{-2em}
    \caption{\textbf{Performance across different maximum iterations}}
    \label{fig:appendix_2}
    \vspace{-1em}
\end{figure}

\noindent\textbf{LoRAs for Multiple NLP tasks Integration.} \textbf{1)} In-domain task integration: We fine-tuned FLAN-t5-large~\cite{https://doi.org/10.48550/arxiv.2210.11416} using LoRA tuning, targeting the \(q\) and \(v\) matrices of all attention layers ~\cite{vaswani2017attention}. The learning rate was set to 5e-5, with a training duration of 10 epochs and a batch size of 12. A warm-up phase \cite{he2016deep} of 100 steps was applied. Optimization was performed using the AdamW optimizer~\cite{kingma2014adam}, with a weight decay of 0.001. \textbf{2)} Multi-task integration: We fine-tuned FLAN-T5-base~\cite{https://doi.org/10.48550/arxiv.2210.11416} using LoRA on the \(q\) and \(v\) matrices across all attention layers~\cite{vaswani2017attention}. The learning rate was set to either \(5 \times 10^{-5}\) or \(1 \times 10^{-5}\), with training durations of 8 or 10 epochs and batch sizes of 12 or 16. A 100-step warm-up phase~\cite{he2016deep} was applied.   
The AdamW optimizer~\cite{kingma2014adam} was used with a weight decay of 0.001. For further detailed training configurations for each subset, please refer to the forthcoming release of our code. 
Table~\ref{tab:performance_scores_multi} and Table~\ref{tab:performance_scores_indomain} summarize the key metrics of the LoRAs trained in this part.

\begin{figure*}
    \centering
    \includegraphics[width=\textwidth]{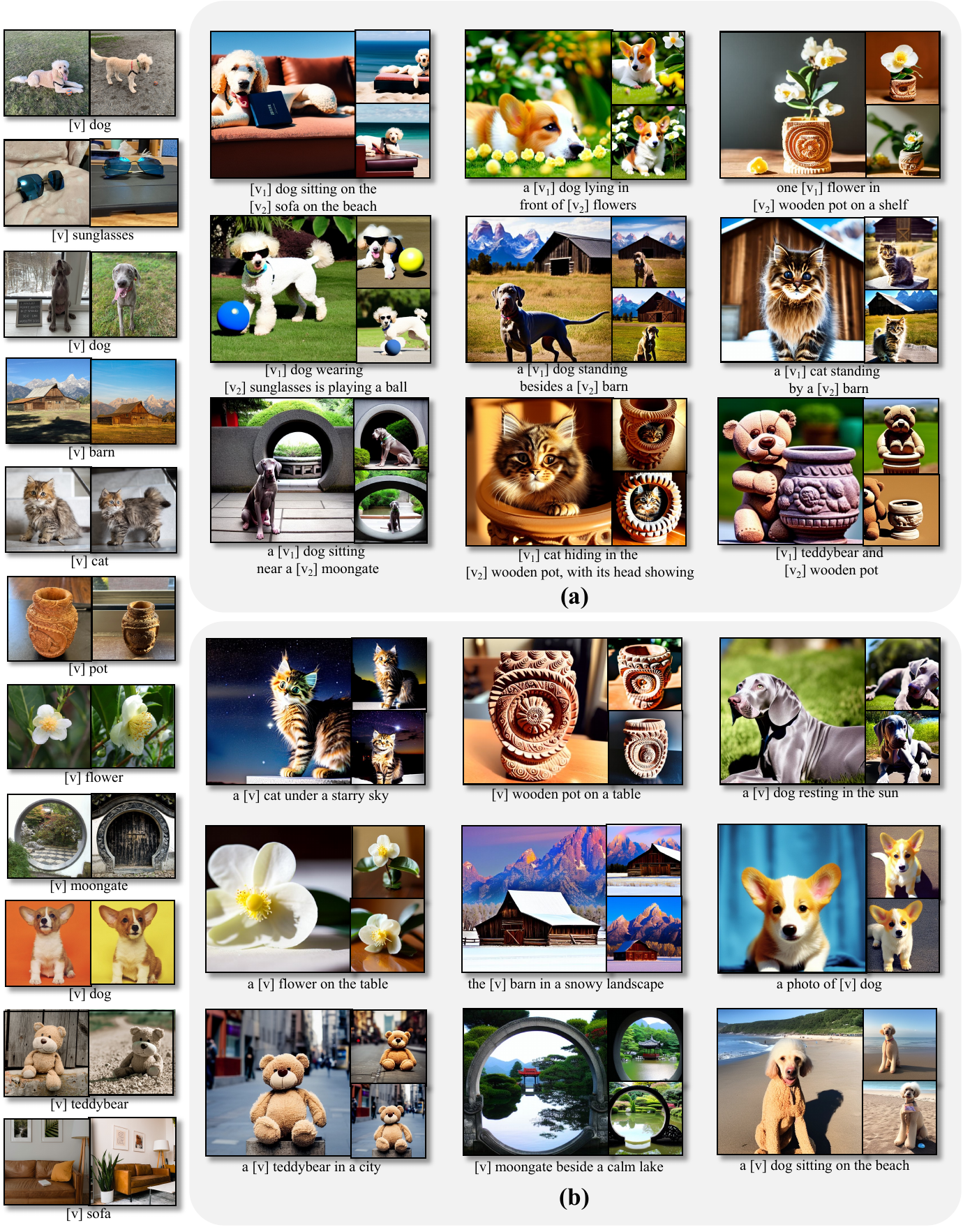} 
    \vspace{-2em}
    \caption{\textbf{Additional qualitative results for multi-concept customization.} Target images represent individual concepts used in the compositions. (a) Examples of pairwise compositions generated by our method. (b) Examples of individual concepts generated by IterIS.}
    \label{fig:appendix_4}
\end{figure*}

\subsection{Baseline Training Details}
\noindent\textbf{Textual Inversion~\cite{gal2022image}.} We employed uniformly initialized learnable tokens with a batch size of 1 and set the gradient accumulation steps to 4. The learning rate was configured to \( 5 \times 10^{-4} \), and the training was carried out over a total of 800 steps. Each new concept was represented as ``[adj\_n] n,'' where ``[adj\_n]'' denoted a learnable token. For instance, when introducing the new concept ``cat,'' we represented the new concept as ``[adj\_cat] cat''.

\noindent\textbf{Custom Diffusion~\cite{kumari2023multi}.} We utilized the official implementation of custom diffusion. To ensure a fair comparison, we adopted its optimization method as the baseline. The number of DDPM steps was set to 200, and the unconditional guidance scale was set to 6.

\noindent\textbf{Linear Merging~\cite{zhang2023composing}.} We performed linear merging using LoRAs with identical weights (e.g., average merging).

\noindent\textbf{RegMean~\cite{jin2022dataless}.} We utilized the official implementation of RegMean, with modifications to the interface. In our experiments, we reproduced RegMean by adopting its official regularization settings, with the regularization coefficient \(\alpha\) set to 0.1. Following the recommendations in the original paper, we set the number of inference samples to 100–200, with a batch size of 16. In cases where the dataset contained fewer than the recommended number, we used all available data. All other hyperparameters and LoRAs were kept consistent with those in our method.
\begin{figure*}
    \centering
    \includegraphics[width=\textwidth]{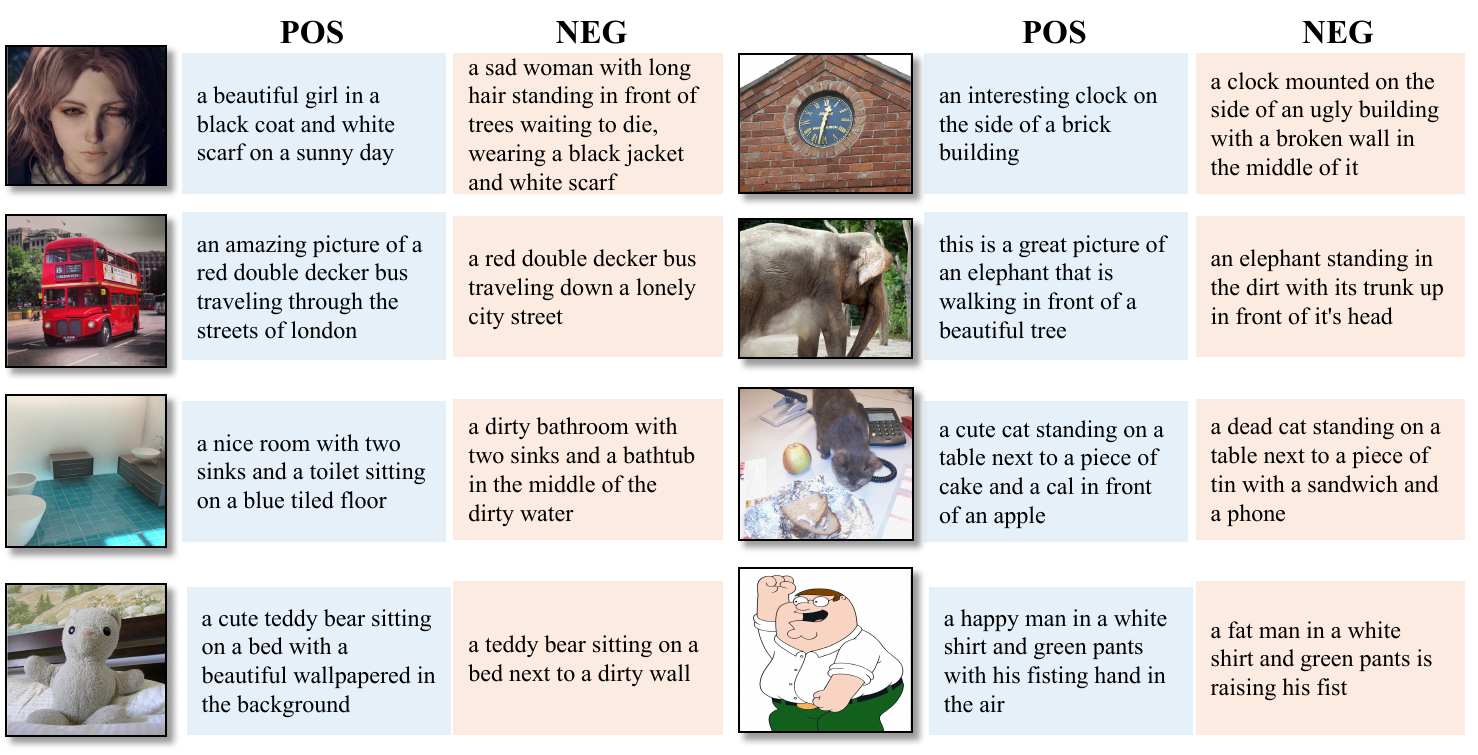} 
    \vspace{-1.5em}
    \caption{\textbf{More examples of style caption generated by our algorithm.}}
    \label{fig:appendix_3}
\end{figure*}

\section{Ablation Study of IterIS}
\label{sec:C}
\noindent\textbf{Influence on Maximum Iteration.} As shown in Figure~\ref{fig:appendix_2} and Table~\ref{tab:iteration_scores}, we evaluated the impact of varying maximum iterations in IterIS on vision-language model. Mean ACC, CIDEr, and score$_{aver}$ (defined as (ACC$_{mean}$ + CIDEr$_{mean}$) / 2) were computed on the validation sets of the POS and NEG datasets. When limited to 4 or fewer iterations, IterIS exhibits notable instability, starting with a low score$_{aver}$ that progressively improves. Peak performance is observed at 5 iterations, beyond which a slight decline suggests potential overfitting. Performance stabilizes after 11 iterations, aligning with our theoretical expectations.
\begin{table}[t]
\centering
\fontsize{9}{10}\selectfont
\renewcommand{\arraystretch}{1.3}
\setlength{\tabcolsep}{4pt} 
\resizebox{\columnwidth}{!}{ 
\begin{tabular}{l|cccccc}
\toprule
    & B-1   & B-2   & B-3   & B-4   & Rough-L & CIDEr \\ \midrule
POS (Ours w/o Reg)  & 0.480 & 0.291 & 0.180 & 0.112 & 0.368   & 0.534 \\
POS (Ours w/ Reg)    & \textbf{0.561} & \textbf{0.359} & \textbf{0.232} & \textbf{0.151} & \textbf{0.420}   & \textbf{0.794} \\
 \midrule
NEG (Ours w/o Reg)  & 0.356 & 0.206 & 0.127 & 0.079 & 0.334   & 0.476 \\ 
NEG (Ours w/ Reg)    & \textbf{0.540} & \textbf{0.348} & \textbf{0.230} & \textbf{0.153} & \textbf{0.406}   & \textbf{0.794} \\
\bottomrule
\end{tabular}}
\caption{\textbf{Performance comparison of our algorithm with and without the regularization term for vision-language model}. The \textbf{bold} numbers highlight the best performance.}
\label{tab:Regularization}
\end{table}

\begin{table}[t] 
\centering
\fontsize{8}{9}\selectfont
\renewcommand{\arraystretch}{1}  
\setlength{\tabcolsep}{3pt}        
\begin{tabular}{l|ccc}
\toprule
Composition & \textbf{(MRPC, SST2)} & \textbf{(COLA, MNLI)} & \textbf{(RTE, MRPC)} \\ \midrule
Ours w/o Reg & (0.272, 0.177) & (0.0, 0.0) &  (0.671, 0.667)\\
Ours w/ Reg & \textbf{(0.824, 0.951)} & \textbf{(0.299, 0.780)} &\textbf{(0.805, 0.814)} \\
\bottomrule
\end{tabular}

\caption{\textbf{Performance comparison of our algorithm with and without the regularization term for LLM.} The first dataset pair consists of MRPC and SST2, and so on. The \textbf{bold} numbers indicate the best performance within each dataset pair.}
\label{tab:regularization_comparison}
\end{table}

\begin{table}[t]
\centering
\fontsize{8}{9}\selectfont  
\renewcommand{\arraystretch}{1.1}  
\setlength{\tabcolsep}{3pt}  
\begin{tabular}{l|ccc}
\toprule
  Composition      & \textbf{(COLA, MNLI)}   & \textbf{(MNLI, SST2)} & \textbf{(COLA, SST2)}  \\
\midrule
Ours w/o Weights & (0.271, 0.728)          & (0.761, 0.945)   & (0.356, 0.943)       \\
Ours w/ Weights   & \textbf{(0.299, 0.780)}          & \textbf{(0.764, 0.945)}    & \textbf{(0.360, 0.943)}       \\
\bottomrule
\end{tabular}
\caption{\textbf{Performance comparison of our algorithm with and without adaptive weights for LLM.} The first dataset pair consists of COLA and MNLI, and so on. The \textbf{bold} numbers indicate the best performance within each dataset pair.}
\label{tab:coef_performance}
\vspace{-2em}
\end{table}

\noindent\textbf{Influence on Regularization Term.}
Using 50 samples for inference, we evaluated the effect of the introduced regularization term on the performance of IterIS for both the vision-language model and the large language model. As demonstrated in Table~\ref{tab:Regularization} and Table~\ref{tab:regularization_comparison}, incorporating the regularization term results in substantial performance improvements, highlighting its pivotal role in enhancing the effectiveness of IterIS.

\noindent\textbf{Influence on Adaptive Weights.}
We assessed the impact of adaptive weights on algorithm performance, as summarized in Table~\ref{tab:coef_performance}. The results demonstrate that adaptive weights can enhance algorithm performance to a certain extent. Moreover, we observe that when there are greater discrepancies between different tasks, the application of our proposed adaptive weights leads to a more noticeable improvement (\eg, COLA and MNLI).

\begin{table}[t]
\centering
\fontsize{7}{8}\selectfont
\renewcommand{\arraystretch}{1}
\setlength{\tabcolsep}{5pt} 
\resizebox{\columnwidth}{!}{%
\begin{tabular}{l|l|cccccc}
\toprule
\textbf{Style}       & \textbf{Method} & \textbf{B-1} & \textbf{B-2} & \textbf{B-3} & \textbf{B-4} & \textbf{Cider} & \textbf{ACC} \\ \midrule
\multirow{3}{*}{\textbf{POS}} 
                     & RegMean         & 0.576 & 0.368 & 0.239 & 0.156 & \textbf{0.801} & 0.624 \\
                     & Linear          & 0.561 & 0.357 & 0.230 & 0.151 & 0.771 & 0.522 \\ 
                     & IterIS          & 0.561 & 0.359 & 0.232 & 0.151 & 0.794 & \textbf{0.830} \\ \midrule
\multirow{3}{*}{\textbf{NEG}} 
                     & RegMean         & 0.540 & 0.341 & 0.223 & 0.144 & 0.779 & 0.692 \\
                     & Linear          & 0.509 & 0.319 & 0.206 & 0.133 & 0.733 & 0.557 \\
                     & IterIS          & 0.540 & 0.348 & 0.230 & 0.153 & \textbf{0.794} & \textbf{0.781} \\ \midrule
\multirow{3}{*}{\textbf{HUM}} 
                     & RegMean         & 0.303 & 0.174 & 0.103 & 0.061 & 0.528 & -\\
                     & Linear          & 0.281 & 0.161 & 0.095 & 0.055 & 0.504 & -\\ 
                    & IterIS          & 0.300 & 0.174 & 0.103 & 0.060 & \textbf{0.530} & -\\ \midrule
\multirow{3}{*}{\textbf{ROM}} 
                     & RegMean         & 0.307 & 0.172 & 0.101 & 0.061 & 0.523 & -\\
                     & Linear          & 0.266 & 0.151 & 0.054 & 0.047 & 0.487 & -\\ 
                     & IterIS          & 0.308 & 0.172 & 0.101 & 0.062 & \textbf{0.524} & -\\ \bottomrule
\end{tabular}%
}
\caption{\textbf{Detailed performance comparison for multi-style caption generation} includes two combinations: \textbf{(1)} ``positive'' (POS) + ``negative'' (NEG), and \textbf{(2)} ``humor'' (HUM) + ``romance'' (ROM). The \textbf{bold} values highlight the best performance. }
\label{tab:style_metrics}
\vspace{-1.5em}
\end{table}

\section{More Results}
\label{sec:D}
\noindent\textbf{More Results on Text-to-Image Diffusion.}
Figure~\ref{fig:appendix_4}(a) showcases additional combinations of pairwise concepts from our experiments, while Figure~\ref{fig:appendix_4}(b) illustrates the individual concepts generated by the composed LoRAs. Our method exhibits robust performance across a wide range of multi-concept combinations.   
\begin{table*}[t]
\centering
\fontsize{8.1}{10}\selectfont
\renewcommand{\arraystretch}{1.2} 
\setlength{\tabcolsep}{3pt} 
\begin{tabular}{l|ccc|ccc|ccc|ccc}
\toprule
                     & \multicolumn{3}{c|}{\textbf{Score$_1$}}      & \multicolumn{3}{c|}{\textbf{Score$_2$}}      & \multicolumn{3}{c|}{\textbf{Score$_3$}}      & \multicolumn{3}{c}{\textbf{Score$_4$}}      \\ 
\cmidrule(lr){2-4} \cmidrule(lr){5-7} \cmidrule(lr){8-10} \cmidrule(lr){11-13}
\textbf{Composition}     & \textbf{IterIS} & \textbf{RegMean} & \textbf{Linear} & \textbf{IterIS} & \textbf{RegMean} & \textbf{Linear} & \textbf{IterIS} & \textbf{RegMean} & \textbf{Linear} & \textbf{IterIS} & \textbf{RegMean} & \textbf{Linear} \\ 
\midrule
Emoint/EC            & \textbf{{0.810}} & 0.808 & 0.784 & \textbf{{0.938}} & 0.922 & 0.912 & -     & -     & -     & -     & -     & -     \\
Emoint/TEC           & \textbf{{0.807}} & 0.758 & 0.731 & \textbf{{0.551}} & 0.544 & 0.474 & -     & -     & -     & -     & -     & -     \\
Emoint/ISEAR         & \textbf{{0.811}} & 0.786 & 0.733 & \textbf{{0.665}} & 0.650 & 0.603 & -     & -     & -     & -     & -     & -     \\
EC/TEC               & \textbf{{0.945}} & 0.926 & 0.881 & 0.603 & \textbf{{0.630}} & 0.592 & -     & -     & -     & -     & -     & -     \\
EC/ISEAR             & \textbf{{0.953}} & 0.938 & 0.900 & 0.709 & \textbf{{0.719}} & 0.679 & -     & -     & -     & -     & -     & -     \\
TEC/ISEAR            & 0.574 & \textbf{{0.592}} & 0.526 & \textbf{{0.710}} & 0.661 & 0.631 & -     & -     & -     & -     & -     & -     \\
Emoint/EC/TEC        & \textbf{{0.782}} & 0.753 & 0.727 & \textbf{{0.942}} & 0.923 & 0.877 & \textbf{{0.540}} & 0.510 & 0.460 & -     & -     & -     \\
Emoint/EC/ISEAR      & \textbf{{0.795}} & 0.774 & 0.720 & \textbf{{0.935}} & 0.926 & 0.887 & \textbf{{0.670}} & 0.641 & 0.593 & -     & -     & -     \\
Emoint/TEC/ISEAR     & \textbf{{0.786}} & 0.721 & 0.683 & \textbf{{0.500}} & 0.499 & 0.455 & \textbf{{0.646}} & 0.607 & 0.593 & -     & -     & -     \\
EC/TEC/ISEAR         & \textbf{{0.950}} & 0.910 & 0.857 & 0.552 & \textbf{{0.555}} & 0.486 & \textbf{{0.683}} & 0.649 & 0.614 & -     & -     & -     \\
Emoint/EC/TEC/ISEAR  & \textbf{{0.776}} & 0.714 & 0.688 & \textbf{{0.935}} & 0.901 & 0.873 & \textbf{{0.508}} & 0.481 & 0.445 & \textbf{{0.657}} & 0.607 & 0.590 \\
\bottomrule
\end{tabular}
\caption{\textbf{Performance of all possible compositions in in-domain task integration.} Score$_i$ denotes the performance metric for the i-th task within the composition. The \textbf{bold} values highlight the best performance.}
\label{tab:results_indomain_task}
\end{table*}
\begin{table}[t]
\centering
\fontsize{8.1}{10}\selectfont
\renewcommand{\arraystretch}{1.2} 
\setlength{\tabcolsep}{2pt} 
\begin{tabular}{l|ccc|ccc}
\toprule
                           & \multicolumn{3}{c|}{\textbf{Score$_1$}}                                                          & \multicolumn{3}{c}{\textbf{Score$_2$}}                                                          \\ \midrule
\textbf{Composition} & \textbf{IterIS}                & \textbf{RegMean}             & \textbf{Linear}              & \textbf{IterIS}                & \textbf{RegMean}             & \textbf{Linear}              \\ \midrule
MNLI/RTE                   & { \textbf{0.831}} & 0.803                        & 0.751                        & { \textbf{0.794}} & 0.787                        & 0.776                        \\
MNLI/COLA                  & { \textbf{0.780}} & 0.755                        & 0.765                        & 0.299                        & 0.279                        & { \textbf{0.383}} \\
MNLI/SST2                  & { \textbf{0.764}} & 0.747                        & 0.645                        & { \textbf{0.945}} & { \textbf{0.945}} & 0.942                        \\
MNLI/QQP                   & { \textbf{0.821}} & 0.813                        & 0.687                        & { \textbf{0.855}} & 0.853                        & 0.842                        \\
MNLI/QNLI                  & { \textbf{0.821}} & 0.806                        & 0.797                        & { \textbf{0.916}} & 0.914                        & 0.905                        \\
MNLI/MRPC                  & { \textbf{0.825}} & 0.816                        & 0.744                        & { \textbf{0.821}} & 0.792                        & 0.772                        \\
RTE/COLA                   & { \textbf{0.787}} & 0.783                        & 0.733                        & 0.311                        & 0.307                        & { \textbf{0.388}} \\
RTE/SST2                   & { \textbf{0.819}} & 0.809                        & 0.773                        & 0.946                        & { \textbf{0.948}} & 0.943                        \\
RTE/QQP                    & 0.816                        & 0.812                        & { \textbf{0.819}} & 0.856                        & { \textbf{0.856}} & 0.853                        \\
RTE/QNLI                   & 0.805                        & 0.805                        & { \textbf{0.812}} & 0.919                        & { \textbf{0.920}} & 0.913                        \\
RTE/MRPC                   & 0.805                        & { \textbf{0.812}} & 0.805                        & { \textbf{0.814}} & 0.801                        & 0.757                        \\
COLA/SST2                  & { \textbf{0.360}} & 0.291                        & 0.233                        & { \textbf{0.943}} & 0.938                        & 0.891                        \\
COLA/QQP                   & 0.386                        & 0.332                        & { \textbf{0.397}} & { \textbf{0.839}} & 0.831                        & 0.734                        \\
COLA/QNLI                  & 0.297                        & 0.305                        & { \textbf{0.346}} & { \textbf{0.906}} & 0.904                        & 0.818                        \\
COLA/MRPC                  & 0.341                        & 0.307                        & { \textbf{0.383}} & 0.811                        & { \textbf{0.816}} & 0.765                        \\
SST2/QQP                   & { \textbf{0.947}} & { \textbf{0.947}} & 0.936                        & { \textbf{0.855}} & 0.854                        & 0.839                        \\
SST2/QNLI                  & { \textbf{0.943}} & { \textbf{0.943}} & { \textbf{0.943}} & { \textbf{0.920}} & 0.916                        & 0.882                        \\
SST2/MRPC                  & { \textbf{0.951}} & 0.947                        & 0.939                        & { \textbf{0.824}} & 0.809                        & 0.794                        \\
QQP/QNLI                   & { \textbf{0.856}} & 0.856                        & 0.846                        & { \textbf{0.920}} & 0.920                        & 0.915                        \\
QQP/MRPC                   & { \textbf{0.855}} & 0.854                        & 0.846                        & { \textbf{0.816}} & 0.814                        & 0.801                        \\
QNLI/MRPC                  & { \textbf{0.922}} & 0.920                        & 0.907                        & { \textbf{0.821}} & { \textbf{0.821}} & 0.797                        \\ \bottomrule
\end{tabular}
\caption{\textbf{Performance of all pairwise compositions in multi-task integration.} Score$_i$ denotes the performance metric for the i-th task within the composition. The \textbf{bold} values highlight the best performance.}
\vspace{-1em}
\label{tab:results_multi_task}
\end{table}

\begin{table}[t] 
\centering
\fontsize{8}{9}\selectfont
\renewcommand{\arraystretch}{1}  
\setlength{\tabcolsep}{3pt}        
\begin{tabular}{l|ccc}
\toprule
\textbf{Composition} & \textbf{(MRPC, SST2)} & \textbf{(SST2, QQP)} & \textbf{(QQP, MRPC)} \\ \midrule
\textbf{Task arithmetic} & (0.806, 0.937) & (0.937, 0.841) &  (0.845, 0.811)\\
\textbf{Ties merging} & (0.765, 0.943) & (0.944, 0.823) &(0.846, 0.806) \\
\textbf{Hyper-linear} & (0.809, 0.933) & (0.931, 0.849) &(0.846, 0.801) \\
\textbf{IterIS (Ours)} & \textbf{(0.824, 0.951)} & \textbf{(0.947,  0.855)} &\textbf{(0.855, 0.816)} \\
\bottomrule
\end{tabular}
\vspace{-1.1em}
\caption{\textbf{Performance for multi-task integration for additional three baselines.}}
\label{tab:NLP_comparison}
\end{table}

\begin{figure}[t]
    \centering
    \includegraphics[width=0.478\textwidth]{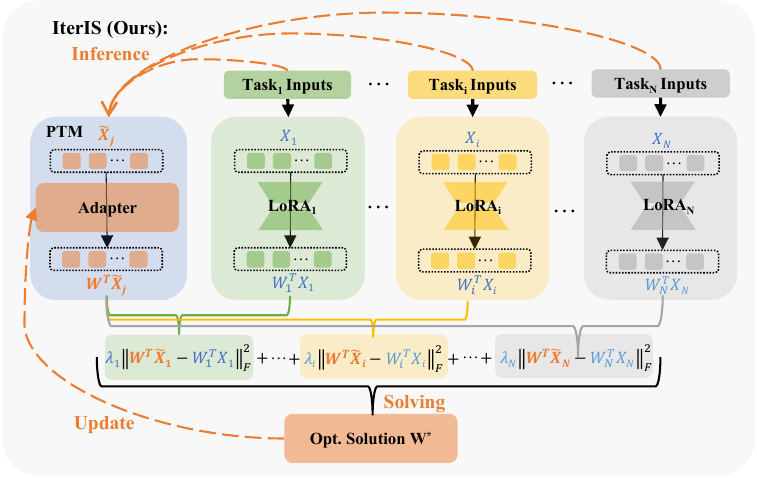} 
    \vspace{-2em}
    \caption{\textbf{The workflow diagram of our algorithm.}}
    \label{fig:appendix_5}
    \vspace{-1em}
\end{figure}
\noindent\textbf{More Results on V\&L Model.} 
As shown in Figure 6 of Section 4, we provide additional examples of positive and negative captions generated by our algorithm. A detailed version of Table 2 of Section 4 from the main text is presented in Table~\ref{tab:style_metrics}. Additionally, we conducted experiments on FlickrStyle10K~\cite{gan2017stylenet}, exploring combinations of two other distinct styles. The results are also included in Table~\ref{tab:style_metrics}. Due to the subjective nature of evaluating humor and romance, as well as the lack of the open test set, we didn't quantify humor(romance)-specific metrics. As a result, aside from the CIDEr score, the advantages of our algorithm for this style combination are less evident. 

\noindent\textbf{More Results on LLM.} 
A detailed version of Table 3 in Section, as discussed in the main text, is provided for reference in this part. Table~\ref{tab:results_indomain_task} summarizes the experimental results of 11 combinations for in-domain task integration, while Table~\ref{tab:results_multi_task} presents the results of 21 combinations for multi-task integration. Our proposed method outperforms both RegMean and linear merging in the majority of cases, highlighting its robustness and effectiveness. Additionally, Table~\ref{tab:NLP_comparison} shows the superiority of our method compared to three extra baselines in the multi-task integration subset.

\section{Others}
\label{sec:E}
\noindent\textbf{Illustration of IterIS.} To facilitate a better understanding of IterIS, we present Figure~\ref{fig:appendix_5}, which effectively illustrates the layer-wise characteristics and inference-solving process of our method.

\noindent\textbf{Efficiency Analysis.} IterIS ensures computation and memory efficiency through sample efficiency and limited iterations. In the V\&L experiments, we utilized 50 prompt-image pairs as inference samples and performed 6 iterations. For each layer, the total computational cost for computing a single inner product matrix is \( 50 \text{ (sample)} \times 6 \text{ (iteration)} = 300 \) one-dimensional tensor inner products. However, that cost for RegMean is \( 1600 \times 1 = 1600 \), which is much higher than IterIS. Moreover, sample efficiency of IterIS allows us to perform all computations without batching, ensuring memory efficiency. Notably, all experiments for IterIS were conducted using a single RTX 3090 GPU.

\noindent\textbf{Detailed Analysis for the Regularization Term and Adaptive Weights.} \textbf{1)} The regularization term prevents irreversible situations in solutions, which often occur in few-sample cases. In such cases, at least one solution to the optimization objective in Eq.(5) renders it zero. This means the output features in each layer of the merged model are equal to those in the corresponding layer of each individual model. Due to the limited number of inference samples, the risk of overfitting to these samples is high. Notably, we can infer that the optimal solution to Eq.(1) is close to the linear solution based on Eq.(2). Applying identity matrices as regularization terms aligns the solution more closely with the linear solution, enhancing robustness. However, diagonal matrices in RegMean cannot achieve the same effect. \textbf{2)} From a dimensional analysis perspective, the scale of the adaptive weights corresponds to that of $\|\bm{X}_{i}\|_F^{-2}$. This neutralizes the scale of $\bm{X}_{i}$ in the optimization objective ($\|\bm{W}_{i}^{T}\bm{X}_{i} - \bm{W}^{T}\tilde{\bm{X}}_{i}  \|_{F}^{2}$), thereby ensuring that the optimization objective remains unbiased to the scale of $\bm{X}_{i}$.

\noindent\textbf{Limitations.} Similar to other LoRA merging methods, performance degradation on individual tasks is inevitable when merging tasks of different types. This degradation becomes more pronounced as the diversity of task types increases or when parameter conflicts arise. Furthermore, since our method does not include domain-specific enhancements for text-to-image diffusion, it shares a limitation with custom diffusion—namely, the confusion of object concepts.


\end{document}